\documentclass[10pt,twocolumn,letterpaper]{article}
\usepackage[pagenumbers]{cvpr} 

\definecolor{cvprblue}{rgb}{0.21,0.49,0.74}
\usepackage[pagebackref,breaklinks,colorlinks,allcolors=cvprblue]{hyperref}

\usepackage[utf8]{inputenc} 
\usepackage[T1]{fontenc}    
\usepackage{hyperref}       
\usepackage{url}            
\usepackage{booktabs}       
\usepackage{amsfonts}       
\usepackage{nicefrac}       
\usepackage{microtype}      
\usepackage{xcolor}         
\usepackage{subfiles}
\usepackage{colortbl}
\usepackage{graphicx}     
\usepackage[skip=3pt,font=small]{subcaption}
\usepackage[skip=3pt,font=small]{caption}
\usepackage{bm}   
\usepackage{acronym}
\usepackage{marvosym}

\usepackage{array}
\usepackage{multirow}

\newlength\savewidth

\usepackage{xspace}

\usepackage{caption} 
\makeatletter
\newcommand{\thickhline}{%
    \noalign {\ifnum 0=`}\fi \hrule height 1pt
    \futurelet \reserved@a \@xhline
}
\newcolumntype{"}{@{\vrule width 1pt}}

\definecolor{mygray}{gray}{.95}
\definecolor{mylightergray}{gray}{.99}
\definecolor{mygreen}{RGB}{10, 179, 33}
\definecolor{mypurple}{HTML}{5E40ED}
\definecolor{myyellow}{HTML}{D5D51A}
\definecolor{myred}{HTML}{DC3C2A}
\definecolor{myblue}{HTML}{519CE8}
\definecolor{myrose}{HTML}{F36DD7}

\definecolor{best}{rgb}{1, 0.8, 0.8}      
\definecolor{second}{rgb}{1, 0.9, 0.6}    
\definecolor{third}{rgb}{1, 1, 0.7}       

\usepackage{pifont}

\usepackage[capitalize]{cleveref}
\crefname{section}{Sec.}{Secs.}
\Crefname{section}{Section}{Sections}
\Crefname{table}{Table}{Tables}
\crefname{table}{Tab.}{Tabs.}

\makeatletter
\DeclareRobustCommand\onedot{\futurelet\@let@token\@onedot}
\def\@onedot{\ifx\@let@token.\else.\null\fi\xspace}
\def\eg{\emph{e.g}\onedot} 

\def\ie{\emph{i.e}\onedot}

\def\etc{\emph{etc}\onedot}

\makeatother

\title{\projname: Real-Time Multi-Person 3D Mesh Estimation via \\ Scale-Adaptive Tokens}

\author{
Chi Su\textsuperscript{1} \quad Xiaoxuan Ma\textsuperscript{1,\Letter} \quad Jiajun Su\textsuperscript{5} \quad Yizhou Wang\textsuperscript{1,2,3,4,\Letter} \\
\textsuperscript{1~}Center on Frontiers of Computing Studies,
School of Computer Science, Peking University \\
\textsuperscript{2~}Inst. for Artificial Intelligence, Peking University \,
\textsuperscript{3~}Nat'l Eng. Research Center of Visual Technology \\ 
\textsuperscript{4~}State Key Laboratory of General Artificial Intelligence, Peking University \\
\textsuperscript{5~}International Digital Economy Academy (IDEA)\\
{\tt\small suchi@stu.pku.edu.cn, \{maxiaoxuan, yizhou.wang\}@pku.edu.cn, sujiajun@idea.edu.cn}\\
{\tt\small $^{\textrm{\Letter}}$ Corresponding authors}} 

\begin{document}

\newcommand{\xiaoxuan}[1]{\textcolor{purple}{\textbf{\small [#1 --xiaoxuan]}}}

\acrodef{sota}[SOTA]{state-of-the-art}
\acrodef{hme}[HME]{human mesh estimation}
\acrodef{vit}[ViTs]{Vision Transformers}
\acrodef{mve}[MVE]{Mean Vertex Error}
\acrodef{mpjpe}[MPJPE]{Mean Per-Joint Position Error}
\acrodef{pa}[PA]{Procrustes-Alignment}
\acrodef{nmve}[NMVE]{Normalized Mean Vertex Error}
\acrodef{nmje}[NMJE]{Normalized Mean Joint Error}
\acrodef{pck}[PCK]{Percentage of Correct Keypoints}
\acrodef{macs}[MACs]{Multiply-Add Cumulation}
\acrodef{gt}[GT]{ground-truth}
\acrodef{gpu}[GPU]{Graphics Processing Unit}
\acrodef{pck}[PCK]{the Percentage of Correctly estimated Keypoints}
\acrodef{mlp}[MLP]{Multi-Layer Perceptron}
\acrodef{fov}[FOV]{field of view}
\acrodef{mae}[MAE]{Mean Absolute Error}

\newcommand{\image}{\mathbf{I}_\text{hr}}
\newcommand{\imgh}{H_\text{hr}}
\newcommand{\imgw}{W_\text{hr}}
\newcommand{\imgsize}{S_\text{hr}}
\newcommand{\lrimage}{\mathbf{I}}
\newcommand{\lrimgh}{H}
\newcommand{\lrimgw}{W}
\newcommand{\lrimgsize}{S}

\newcommand{\camint}{\bm{K}}
\newcommand{\focal}{f}
\newcommand{\gtfocal}{\tilde{f}}
\newcommand{\principalu}{p_\text{u}}
\newcommand{\principalv}{p_\text{v}}

\newcommand{\patchsize}{P}
\newcommand{\featuredim}{D}

\newcommand{\tokens}{\mathcal{T}}
\newcommand{\token}{t}
\newcommand{\numtokens}{k}

\newcommand{\lrtokens}{\mathcal{T}_\text{LR}}
\newcommand{\lrtoken}{t^\text{lr}}
\newcommand{\numlrtokens}{k_\text{lr}}

\newcommand{\smalltokens}{\mathcal{T}_\text{SMALL}}
\newcommand{\smalltoken}{t^{\text{small}}}
\newcommand{\numsmalltokens}{k_\text{small}}

\newcommand{\largetokens}{\mathcal{T}_\text{LARGE}}
\newcommand{\largetoken}{t^{\text{large}}}
\newcommand{\numlargetokens}{k_{\text{large}}}

\newcommand{\hrtokens}{\mathcal{T}_\text{HR}}
\newcommand{\hrtoken}{t^\text{hr}}
\newcommand{\numhrtokens}{k_\text{hr}}

\newcommand{\bgtokens}{\mathcal{T}_\text{B}}
\newcommand{\bgtoken}{t^\text{b}}
\newcommand{\numbgtokens}{k_\text{b}}

\newcommand{\bgtokenspooled}{\mathcal{T}'_\text{B}}
\newcommand{\bgtokenpooled}{t^{\text{b}'}}
\newcommand{\numbgtokenspooled}{{k}'_\text{b}}

\newcommand{\mrtokens}{\mathcal{T}_\text{SA}}
\newcommand{\mrtoken}{t^\text{sa}}
\newcommand{\nummrtokens}{k_\text{sa}}

\newcommand{\queries}{\mathcal{Q}}
\newcommand{\query}{q}
\newcommand{\numqueries}{n}

\newcommand{\scalemap}{\mathbf{S}}
\newcommand{\diagbox}{d_\text{bb}}
\newcommand{\scale}{s}
\newcommand{\conf}{c}
\newcommand{\mapthreshscale}{\alpha_\text{s}}
\newcommand{\mapthreshconf}{\alpha_\text{c}}

\newcommand{\smplshape}{\bm{\beta}}
\newcommand{\smplpose}{\bm{\theta}}
\newcommand{\smpl}{\mathcal{M}}
\newcommand{\jointnumber}{J}
\newcommand{\vertsnumber}{6890}
\newcommand{\jregressor}{\mathbf{M}}
\newcommand{\mesh}{\mathbf{V}}
\newcommand{\transl}{\bm{t}}

\newcommand{\encoder}{\mathcal{E}}
\newcommand{\decoder}{\mathcal{D}}
\newcommand{\posehead}{\mathcal{H}_\text{p}}
\newcommand{\shapehead}{\mathcal{H}_\text{s}}
\newcommand{\translhead}{\mathcal{H}_\text{t}}
\newcommand{\boxhead}{\mathcal{H}_\text{b}}
\newcommand{\confhead}{\mathcal{H}_\text{c}}

\newcommand{\numenc}{N_\text{e}}
\newcommand{\numlrenc}{N_\text{lr}}
\newcommand{\numhrenc}{N_\text{hr}}
\newcommand{\nummrenc}{N_\text{sa}}
\newcommand{\numdec}{N_\text{d}}

\newcommand{\loss}{\mathcal{L}}
\newcommand{\lossmap}{\loss_\text{map}}
\newcommand{\lossdepth}{\loss_\text{depth}}
\newcommand{\losspose}{\loss_\text{pose}}
\newcommand{\lossshape}{\loss_\text{shape}}
\newcommand{\lossjddd}{\loss_\text{j3d}}
\newcommand{\lossjdd}{\loss_\text{j2d}}
\newcommand{\lossbox}{\loss_\text{box}}
\newcommand{\lossdet}{\loss_\text{det}}
\newcommand{\weight}{\mathcal{\lambda}}
\newcommand{\weightmap}{\weight_\text{map}}
\newcommand{\weightdepth}{\weight_\text{depth}}
\newcommand{\weightpose}{\weight_\text{pose}}
\newcommand{\weightshape}{\weight_\text{shape}}
\newcommand{\weightjddd}{\weight_\text{j3d}}
\newcommand{\weightjdd}{\weight_\text{j2d}}
\newcommand{\weightbox}{\weight_\text{box}}
\newcommand{\weightdet}{\weight_\text{det}}

\newcommand{\E}{\mathbb{E}}
\newcommand{\R}{\mathbb{R}}

\newcommand{\joint}{\mathbf{J}}  

\newcommand{\gtmesh}{\tilde{\mathbf{V}}}

\newcommand{\threshdet}{\alpha_\text{d}}

\newcommand{\depth}{d}
\newcommand{\gtdepth}{\tilde{d}}

\newcommand{\projpage}{https://ChiSu001.github.io/SAT-HMR/}
\newcommand{\projname}{SAT-HMR}

\twocolumn[{%
    \renewcommand\twocolumn[1][]{#1}%
    \maketitle
    \vspace{-2.2em}
    \begin{center}
        \begin{minipage}[t]{0.52\linewidth}
            \centering
            \includegraphics[height=5.7cm]{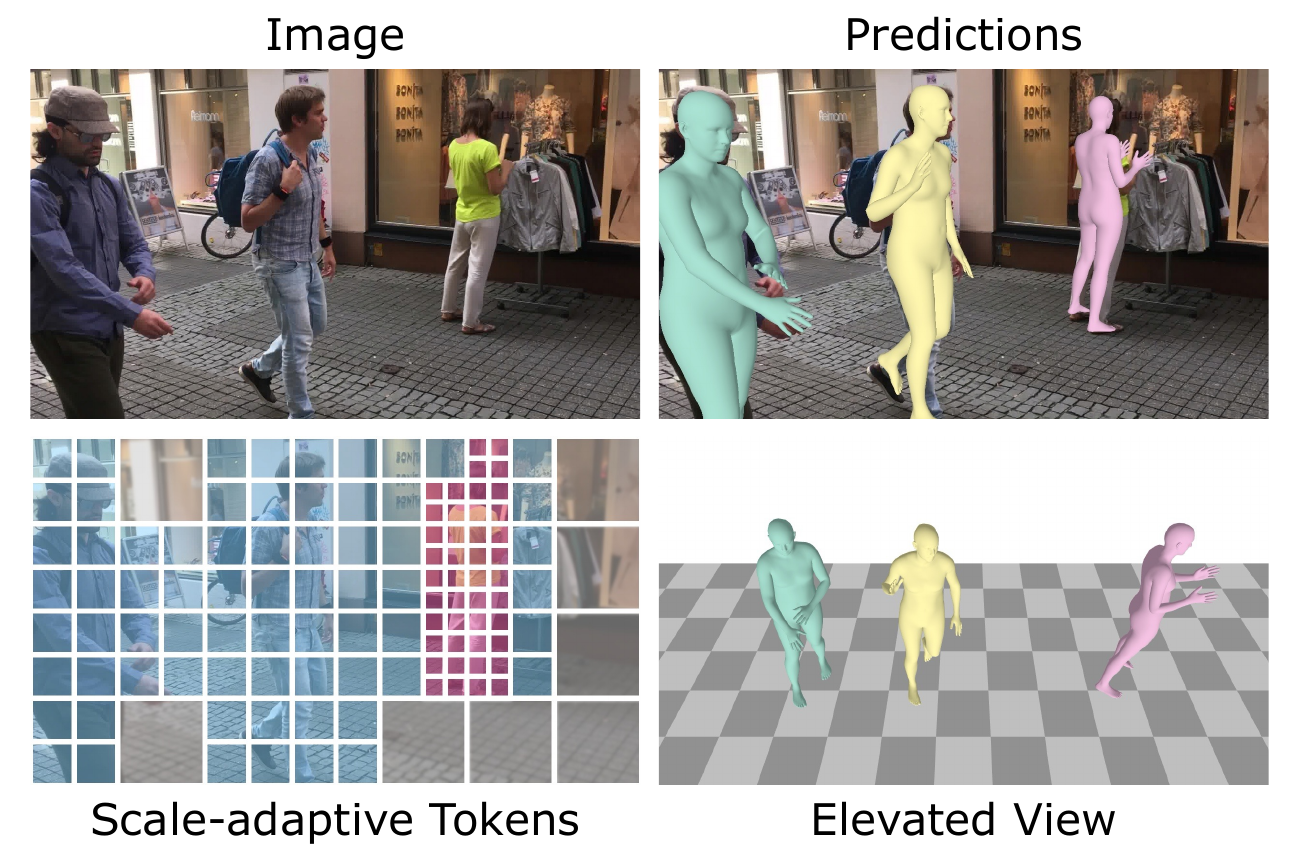}  
            {(a)}  
        \end{minipage}%
        \hfill%
        \begin{minipage}[t]{0.47\linewidth}
            \centering
            \includegraphics[height=5.5cm]{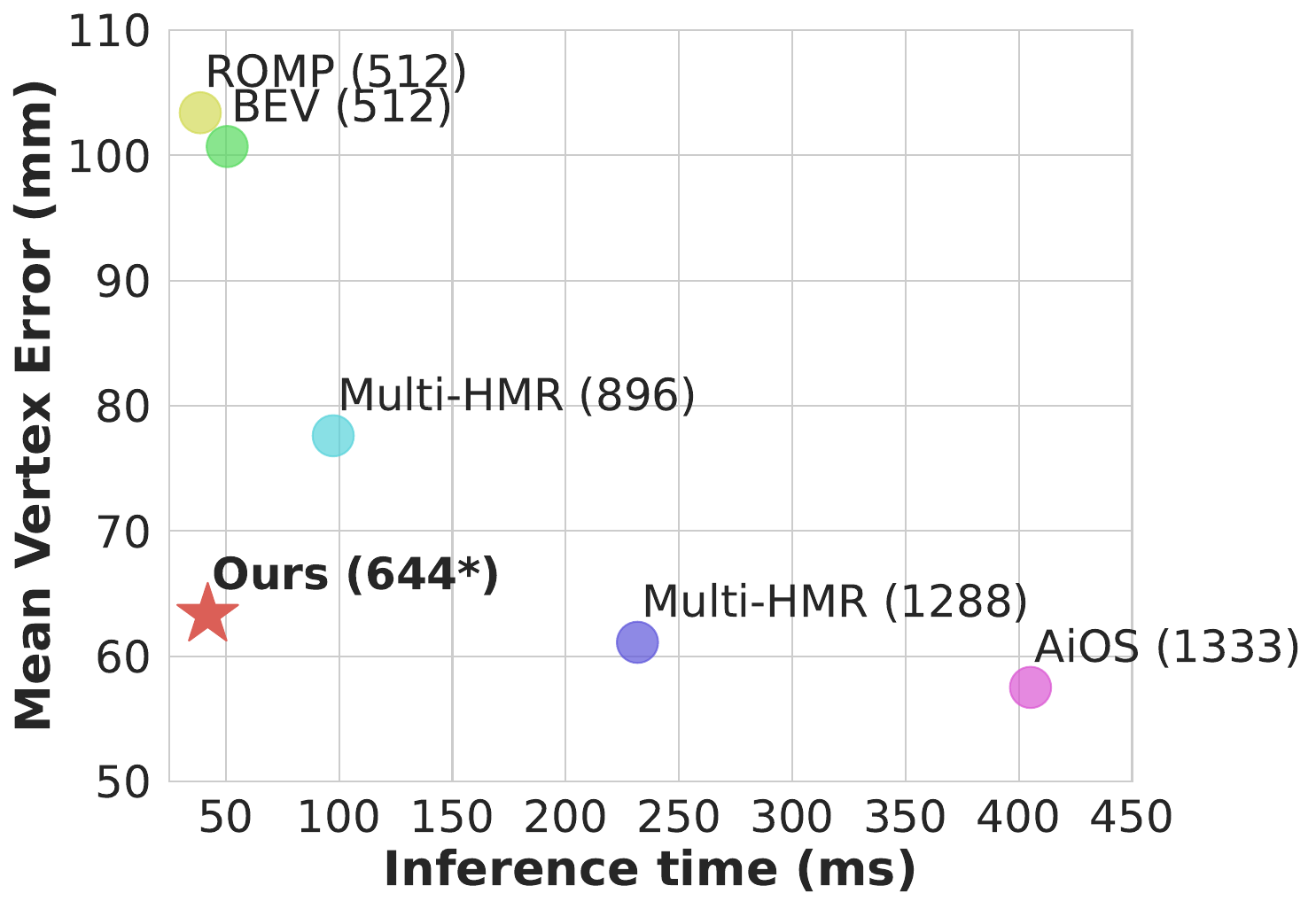}  
            {(b)}  
        \end{minipage}
        \captionof{figure}{\textbf{(a) We propose scale-adaptive tokens in our one-stage framework for real-time multi-person 3D mesh estimation.} Our method introduces scale-adaptive tokens, dynamically adjusted based on the relative size of individuals in the image, to more efficiently encode features, enabling real-time and accurate multi-person mesh estimation. We present a conceptual visualization of the scale-adaptive tokens. The right column visualizes the predicted meshes projected onto an image from 3DPW \cite{vonMarcard2018} dataset and from an elevated view. \textbf{(b) Comparison of estimation error and inference time across different methods, with input resolutions in parentheses.} Our method, using a mixed resolution with a base resolution of 644, achieves comparable performance to state-of-the-art methods on AGORA \cite{patel2021agora} test set while maintaining real-time inference efficiency. Code and models are available at \url{\projpage}.}
        \label{fig:teaser}
    \end{center}
    \vspace*{0.1cm}  
}]

\begin{abstract}
We propose SAT-HMR, a one-stage framework for real-time multi-person 3D human mesh estimation from a single RGB image. While current one-stage methods, which follow a DETR-style pipeline, achieve state-of-the-art (SOTA) performance with high-resolution inputs, we observe that this particularly benefits the estimation of individuals in smaller scales of the image (e.g., those of young age or far from the camera), but at the cost of significantly increased computation overhead. To address this, we introduce scale-adaptive tokens that are dynamically adjusted based on the relative scale of each individual in the image within the DETR framework. Specifically, individuals in smaller scales are processed at higher resolutions, larger ones at lower resolutions, and background regions are further distilled. These scale-adaptive tokens more efficiently encode the image features, facilitating subsequent decoding to regress the human mesh, while allowing the model to allocate computational resources more effectively and focus on more challenging cases. Experiments show that our method preserves the accuracy benefits of high-resolution processing while substantially reducing computational cost, achieving real-time inference with performance comparable to SOTA methods. 
\end{abstract}
\vspace{-0.3cm}

\section{Introduction}

\begin{figure}[t]
  \centering
  \includegraphics[width=0.95\linewidth]{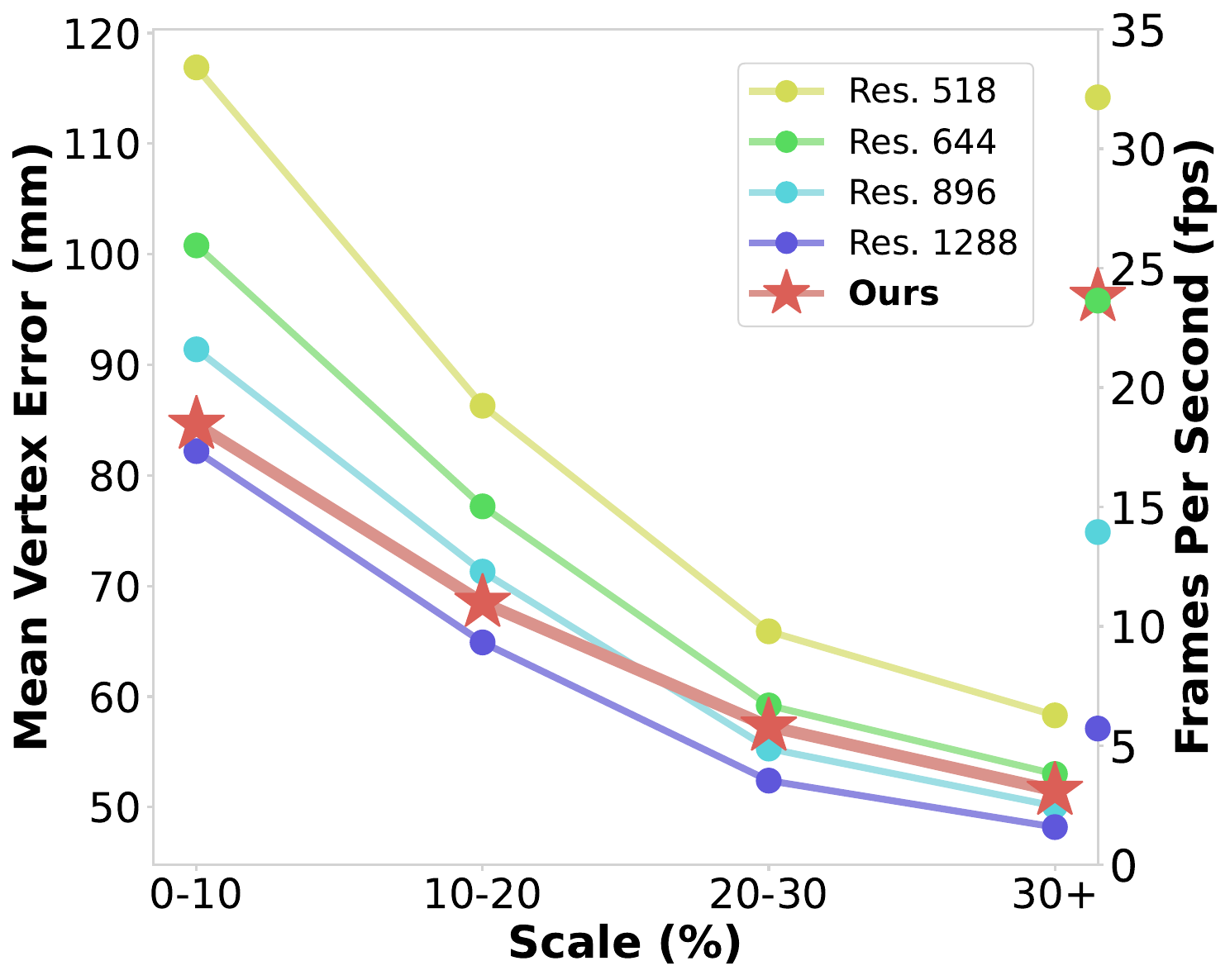}
  \caption{\textbf{Estimation errors and FPS of baselines with different resolutions and our method across individuals at various scales.} The scale of an individual refers to the person's size relative to the overall image and please refer to \cref{subsec:scaleaware} for mathematical definition. The colored lines show the \ac{mve} errors (left y-axis) of the baselines with different resolutions (Res.) on the AGORA \cite{patel2021agora} validation set. The colored markers on the right y-axis indicate the FPS of the corresponding models. Our method adopts a mixed resolution with a base resolution of 644.}
  \label{fig:ablation_scale_fps}
  \vspace{-0.8em}
\end{figure}

Multi-person 3D human mesh estimation from a single RGB image aims to localize all individuals in the scene and estimate their 3D meshes, typically represented by a parametric human model such as SMPL \cite{loper2015smpl}. This is a fundamental task with broad applications, spanning social interaction \cite{salzmann2023robots,zhu2023social}, game production~\cite{zhu2024human}, \etc. Existing methods primarily focus on addressing challenges such as monocular depth ambiguity and inter-person interactions, and can be broadly categorized into two types: multi-stage and one-stage approaches. Multi-stage methods \cite{choi2022learning, qiu2022dynamic, zanfir2018deep, zanfir2018monocular, jiang2020coherent} typically detect and process each individual by cropping and resizing them to a uniform, relatively high resolution, which generally improves performance. However, this overlooks the global context, such as the original scales of individuals and their relative positions within the entire image, potentially leading to suboptimal results.

To address these limitations, one-stage methods gradually emerge as a promising alternative \cite{sun2021monocular, sun2022putting, sun2024aios, multihmr2024}. These approaches operate on the entire image without explicit cropping, preserving global context and enabling end-to-end training. Building on the success of DETR \cite{carion2020end} in object detection, recent works \cite{sun2024aios, multihmr2024} adopt a DETR-style pipeline. They mainly focus on designing queries and decoders to estimate the human mesh, \eg Multi-HMR \cite{multihmr2024} uses queries to represent each person and decodes the corresponding human mesh. However, these methods rely on high-resolution inputs to achieve \ac{sota} results, as shown in \cref{fig:teaser} (b), which significantly increases computational cost and limits real-time processing capabilities. 

Building on the findings of \cite{sun2024aios, multihmr2024}, we observe that high-resolution input plays a crucial role in improving accuracy. To explore this, we design a naive baseline following similar architecture and analyze how its estimation error changes for individuals of different scales as we increase the input resolution. We define scale as the size of the person relative to the entire image\footnote{See \cref{eq:scale_def} for detailed definition of scale.}. As shown in \cref{fig:ablation_scale_fps}, increasing the input resolution consistently leads to a decrease in the estimation error, particularly for smaller-scale individuals (\eg, kids, those far from the camera or those in closed-poses). Notably, for individuals in the 0-10\% scale, we observe an error decrease of nearly 35mm when comparing the highest-resolution model (\textcolor{mypurple}{blueviolet}) to the lowest-resolution one (\textcolor{myyellow}{yellow}). However, processing the entire image at high resolution incurs significant computational costs, especially when individuals are close to the camera and occupy a large portion of the image. In such cases, these individuals are already represented by many tokens, and using more tokens with high resolution would only greatly increase the cost while offering marginal improvement (\ie, a slight error reduction in cases where individuals cover 30+\% of the image in \cref{fig:ablation_scale_fps}). Therefore, both our best baseline model (\textcolor{mypurple}{blueviolet}) and recent \ac{sota} methods \cite{sun2024aios,multihmr2024} in \cref{fig:teaser} (b) fail to achieve real-time inference and run at about 5 FPS, highlighting a common limitation. 

To address this, we extend our baseline built on DETR \cite{carion2020end} and shift our focus toward more efficient feature encoding. Specifically, we introduce \textbf{scale-adaptive tokens}, which are dynamically adjusted based on the scale of individuals in the image. We first predict a patch-level scale map, assigning a scale value to each token. Tokens corresponding to smaller-scale individuals are replaced with higher-resolution tokens. Additionally, background tokens identified by the scale map are further distilled to reduce computational cost. These scale-adaptive tokens capture individuals at varying scales with the appropriate level of detail, as shown in \cref{fig:teaser} (a). The tokens are then processed by the decoder and prediction heads to regress the human mesh. By efficiently allocating computational resources to the more challenging cases, our method strikes a better balance between accuracy and speed. As shown in \cref{fig:ablation_scale_fps}, our method (\textcolor{myred}{red}) achieves performance comparable to high-resolution input while maintaining real-time inference speed at 24 FPS. Experiments on benchmark datasets further demonstrate the effectiveness of our method, achieving competitive performance with high efficiency.

To conclude, our contributions are as follows:
\begin{itemize}
    \item We propose scale-adaptive tokens that are dynamically adjusted to handle individuals of varying scales, enabling more efficient feature encoding.
    \item By employing scale-adaptive tokens, our method achieves performance comparable to using full-image high resolution, but with significantly lower computational cost.
    \item Our approach achieves \ac{sota} performance with up to 5x the speed of the top-performing methods, making it the \textbf{best real-time} model for multi-person 3D mesh estimation, running at 24 FPS.
\end{itemize}

\section{Related work}
\subsection{Multi-Person Human Mesh Estimation}
Multi-person human mesh estimation from a single RGB image aims to regress all human meshes, typically represented by parametric models such as SMPL \cite{loper2015smpl} and SMPL-X \cite{pavlakos2019expressive}. Existing methods can be categorized into \textit{multi-stage} and \textit{one-stage} approaches.
Multi-stage methods \cite{goel2023humans, choi2022learning, qiu2022dynamic, zanfir2018deep, zanfir2018monocular, jiang2020coherent, fastervoxelpose} use off-the-shelf human detectors \cite{renNIPS15fasterrcnn} to crop each person's region, followed by single-person mesh estimation \cite{zhang2021pymaf,ma20233d,zhu2023motionbert,Xu_2024_CVPR,ma2025vmarker}. This approach preserves relatively high-resolution images for each cropped region, generally leading to higher accuracy \cite{sun2024aios}. However, its reliance on detected bounding boxes and the lack of global context make it challenging to handle occlusion \cite{sun2021monocular}.

In contrast, one-stage methods \cite{sun2021monocular,sun2022putting,qiu2023psvt,sun2024aios,multihmr2024} estimate human meshes for all individuals by encoding the entire image and predicting simultaneously. Pioneer work like ROMP \cite{sun2021monocular} and BEV \cite{sun2022putting} use CNN backbones \cite{he2016deep} to extract global features for mesh regression. However, the low input resolution limits the expressiveness of the distilled features \cite{sun2024aios}, leading to suboptimal performance. 
Recent methods like AiOS \cite{sun2024aios} and Multi-HMR \cite{multihmr2024} adopt DETR-style architectures to achieve \ac{sota} performance at higher resolutions, but with significant computational overhead, making real-time use impractical (\eg, Multi-HMR runs at only 4 FPS as shown in \cref{fig:teaser}). Instead, we argue that tokenizing the entire image at high resolution is unnecessary. By adjusting patch resolution based on the scale and focusing computational resources on more challenging areas, we achieve real-time inference with competitive performance.

\subsection{Transformers in Human Mesh Estimation}
Transformers \cite{vaswani2017attention}, originally designed for sequential inputs, have demonstrated strong performance in various vision tasks, such as image classification with \ac{vit} \cite{dosovitskiy2020image}. And it has been widely adopted in human mesh estimation \cite{dou2023tore, lin2021end, lin2021mesh, cho2022FastMETRO, yang2023capturing, sun2024aios, multihmr2024}. For single-person settings, methods such as \cite{dou2023tore, lin2021end, lin2021mesh, cho2022FastMETRO, yang2023capturing} focus on improving the accuracy through various architectural designs. 
Inspired by the DETR series \cite{carion2020end,liu2022dabdetr,zhang2022dino} in object detection, recent methods have adopted a DETR-style framework for multi-person pose and mesh estimation \cite{liu2023group, yang2023explicit, shi2022end, sun2024aios, multihmr2024}, enabling one-stage estimation for all individuals. These works primarily focus on designing decoder queries to represent humans, keypoints, and mesh parameters. For instance, AiOS \cite{sun2024aios} refines predictions by progressively filtering and expanding queries in the decoder, while Multi-HMR \cite{multihmr2024} uses each query to represent a detected person. Our method also builds upon the DETR framework but focuses on improving token representations in the encoder to capture scale-aware human features, enabling accurate 3D mesh regression.

\subsection{Efficient Vision Transformers}
Although \ac{vit} \cite{dosovitskiy2020image} achieves significant success across various vision tasks, their quadratic computational complexity remains a major challenge, particularly for tasks that require dense predictions such as object detection \cite{tang2022quadtree}. As a result, efforts to accelerate \ac{vit} primarily focus on reducing the overall number of image tokens. For instance, \cite{rao2021dynamicvit, yin2022vit, meng2022adavit} drop less informative tokens, \cite{bolya2022token, renggli2022learning} merge similar tokens, and \cite{ronen2023vision} proposes a mixed-resolution image tokenization scheme. Recently, TORE \cite{dou2023tore} accelerates single-person 3D mesh estimation by pruning background tokens. However, since background still provides important context, entirely discarding it leads to performance degradation (see \cref{subsec:ablation}). In this work, we propose a new perspective with scale-adaptive tokens that achieve a more effective trade-off between performance and computational cost.

\section{Method}
\subsection{Preliminaries}
\label{subsec:preliminaries}
We use SMPL \cite{loper2015smpl} as our body model $\smpl$, which uses pose and shape parameters to represent the human body. Specifically, the pose parameters $\smplpose \in \R^{24 \times 3}$ are the relative rotations of 24 body joints, and the shape parameters $\smplshape \in \R^{10}$ describe the body shape. The model outputs a human mesh $\mesh = \smpl(\smplpose, \smplshape) \in \R^{\vertsnumber \times 3}$ and the 3D joints can be obtained via $\joint = \jregressor\mesh \in \R^{\jointnumber \times 3}$, where $\jregressor \in \R^{\jointnumber \times \vertsnumber}$ is the joint regressor, and $\jointnumber$ represents the number of joints.

\subsection{Overview}
\label{subsec:overview}

Building on recent methods \cite{sun2024aios, multihmr2024}, we utilize a DETR-style pipeline \cite{carion2020end}, consisting of a Transformer encoder, decoder, and prediction heads for regressing SMPL parameters.

\begin{figure*}[t]
  \centering
  \includegraphics[width=0.96\linewidth]{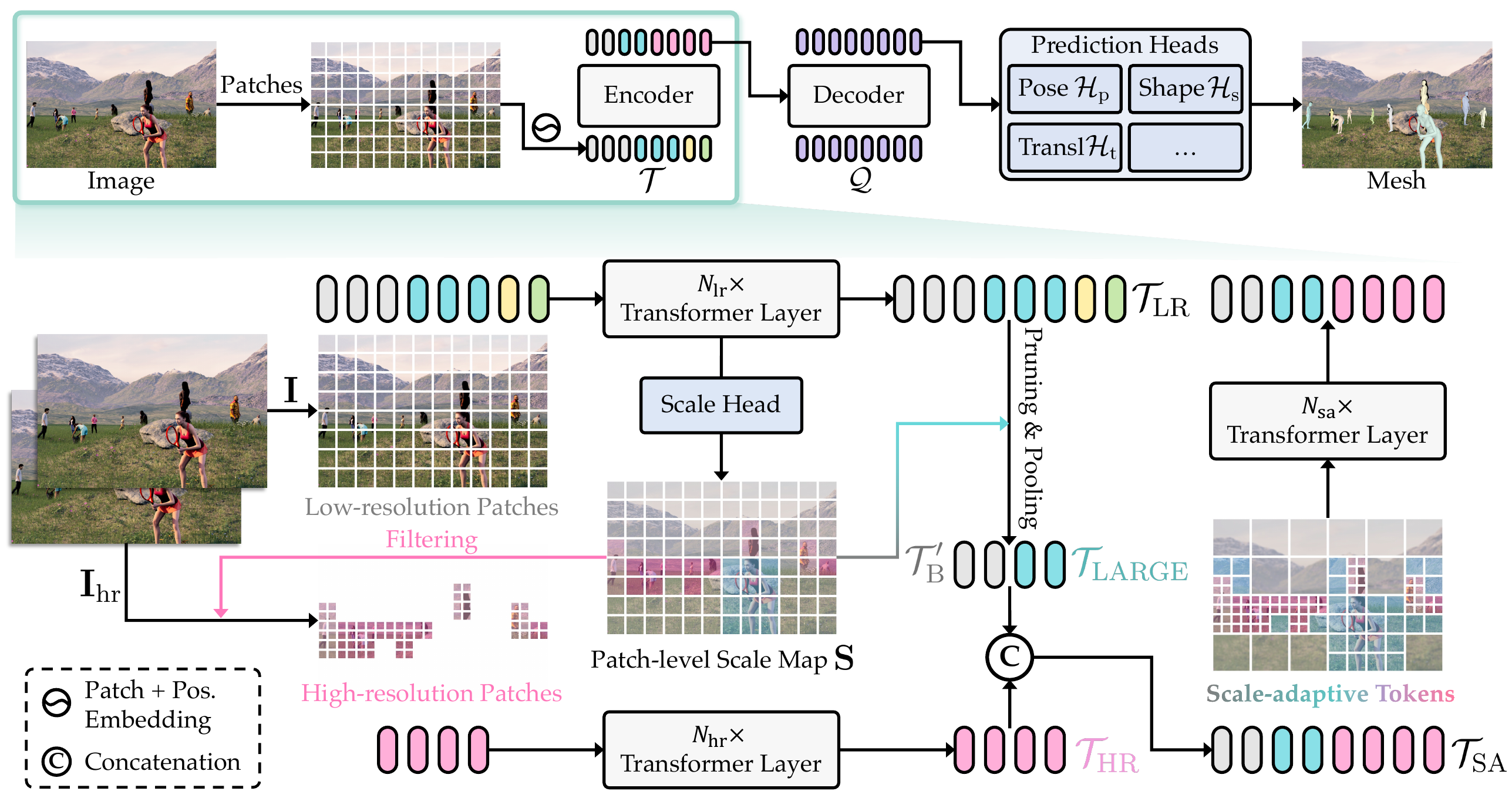}
  \caption{\textbf{Overview of (top) the baseline method and (bottom) our method with scale-adaptive tokens.} \textbf{Top:} Our baseline method adopts a DETR-style \cite{carion2020end} pipeline consisting of a Transformer encoder, decoder, and prediction heads for regressing SMPL parameters. \textbf{Bottom:} Our method focuses on efficient feature encoding using scale-adaptive tokens. Specifically, low-resolution and high-resolution patches are extracted from the input images $\lrimage$ and $\image$, respectively. A scale head network predicts a patch-level scale map $\scalemap$ from the low-resolution tokens, classifying them into three categories: background, small-scale, and large-scale. This scale map guides the pruning and pooling of low-resolution tokens $\lrtokens$ and indicates which patches should be replaced by high-resolution ones. By concatenating the pooled background tokens $\bgtokenspooled$, the remaining large-scale low-resolution tokens $\largetokens$, and the high-resolution tokens $\hrtokens$, we obtain scale-adaptive tokens $\mrtokens$. These tokens are then processed by the encoder, decoder, and multiple prediction heads to regress the human mesh.}
  \label{fig:pipeline}
  \vspace{-0.8em}
\end{figure*}

\vspace{0.4em}
\noindent\textbf{Naive baseline.} While high-resolution inputs are shown to significantly enhance estimation accuracy \cite{sun2024aios, multihmr2024}, they also introduce considerable computational overhead. To further explore the trade-off between resolution and efficiency, we first introduce a simple baseline using a DETR-style architecture, as depicted in \cref{fig:pipeline} (top).

Given an RGB image $\image \in \R^{\imgh \times \imgw \times 3}$, it is divided into regular patches of size $\patchsize \times \patchsize$. Patch embedding and positional embedding are then applied to these image patches, resulting in a sequence of image feature tokens $\tokens = \{\token_1, \token_2, \ldots, \token_{\numtokens}\}$, where $\numtokens = \imgh/\patchsize \times \imgw/\patchsize$. These tokens are subsequently processed by a Transformer encoder.
In the decoder, a set of human queries is initialized as $\queries = \{\query_1, \query_2, \ldots, \query_{\numqueries}\}$, where $\numqueries$ is a predefined hyperparameter large enough to detect all the individuals. 
After being processed by the decoder, these human queries are passed through multiple prediction heads: pose head $\posehead$, shape head $\shapehead$, translation head $\translhead$, and box head $\boxhead$ to predict SMPL parameters (pose $\smplpose$, shape $\smplshape$, and 3D translation $\transl$) and bounding boxes, respectively, Each of the four prediction heads is implemented as an \ac{mlp}. Additionally, a linear projection is applied to the human queries to compute confidence scores, which are used to filter valid predictions with a threshold $\threshdet$.

\vspace{0.4em}
\noindent\textbf{Motivation.} We evaluate the baseline with different image resolutions as input (see \cref{fig:ablation_scale_fps}, \cref{subsec:ablation} for details) and observe that high-resolution inputs significantly improve estimation accuracy, particularly for small-scale individuals whose size is relatively small to the image. However, for large-scale individuals, \eg, those closer to the camera, the benefits of higher resolution are much less pronounced. Despite these improvements, high-resolution inputs bring a substantial increase in computational cost, which is consistent with findings from prior work \cite{sun2024aios, multihmr2024} (see \cref{fig:teaser}).

This motivates us to explore more efficient strategies that balance performance and speed and we aim to encode the image feature more efficiently from an earlier stage. For large-scale individuals, additional tokens may be unnecessary; instead, computational resources should be allocated more effectively. To address this, we introduce \textbf{scale-adaptive tokens}, which replace the uniformly partitioned tokens used in the baseline. In the following sections, we will explain how the scale is defined and computed, and how the tokens are adjusted accordingly (\cref{subsec:scaleaware}), followed by a discussion of the training process (\cref{subsec:loss}).

\subsection{Scale-Adaptive Tokens}
\label{subsec:scaleaware}
As shown in \cref{fig:pipeline} (bottom), we begin with a low-resolution image $\lrimage \in \R^{\lrimgh \times \lrimgw \times 3}$, where $(\lrimgh, \lrimgw) = (\imgh/2, \imgw/2)$, and first uniformly partition it into low-resolution patches. Our goal is to adjust the resolution of each patch based on the scale of the individual it represents. To achieve this, we introduce a \textit{patch-level scale map} that guides the resolution adjustments for each patch. Specifically, this map determines which low-resolution tokens should be further subdivided and replaced by high-resolution ones, and which tokens can be compressed. The high-resolution patches are extracted from a corresponding high-resolution image $\image$. After being processed by the Transformer layers, the remaining low-resolution tokens and the newly generated high-resolution tokens are combined into \textit{scale-adaptive tokens}, which provide a more efficient representation with varying levels of detail. These scale-adaptive tokens are then passed through additional Transformer layers and, finally, to the decoder to regress the SMPL parameters.

\vspace{0.4em}
\noindent\textbf{Patch-level scale map.} We define a patch-level scale map $\scalemap \in [0,1]^{\frac{\lrimgh}{\patchsize} \times \frac{\lrimgw}{\patchsize} \times 2}$ for patches extracted from the low-resolution image $\lrimage$. For each patch, the scale map $\scalemap(i, j) = (\conf, \scale)$ contains two values: $\conf$ represents the confidence indicating whether the patch overlaps with an individual, where 0 denotes the background; $\scale$ represents the scale of the individual within the patch, defined as
\vspace{-0.3em}
\begin{equation} 
\label{eq:scale_def}
\scale = \min(\diagbox/\imgsize, 1), 
\end{equation}
where $\diagbox$ is the diagonal length of the bounding box in $\image$, and $\imgsize = \max(\imgh, \imgw)$. If the patch overlaps with multiple individuals, we use the scale of the individual closest to the camera, \ie, the one with the smaller depth. The scale is only defined for patches overlapping with individuals, and for background patches, the scale value is ignored.

To predict the patch-level scale map, we pass the low-resolution patches through a shallow transformer encoder with $\numlrenc$ layers, followed by a scale head network, implemented as an \ac{mlp}. As illustrated in \cref{fig:pipeline}, patches without color are classified as background, while colored regions represent patches overlapping with individuals. A color gradient from \textcolor{myblue}{blue} to \textcolor{myrose}{rose} indicates the scale of the individuals, with blue representing large-scale individuals and rose representing small-scale individuals.

\vspace{0.4em}
\noindent\textbf{Scale-adaptive encoder.} Building on the predicted scale map $\scalemap$, the low-resolution tokens $\lrtokens = \{\lrtoken_1, \lrtoken_2, \ldots, \lrtoken_{\numlrtokens}\}$ generated from the low-resolution patches are then classified into three groups: background $\bgtokens$, small-scale $\smalltokens$, and large-scale $\largetokens$ tokens. Tokens identified as background, based on a confidence score $\conf$ that meets a threshold $\mapthreshconf$, are marked as background tokens $\bgtokens = \{\bgtoken_1, \bgtoken_2, \ldots, \bgtoken_{\numbgtokens}\}$. The remaining tokens are further categorized by their scale values $\scale$ using a scale threshold $\mapthreshscale$ and are assigned to either the \textit{small-scale} category, denoted by $\smalltokens = \{\smalltoken_1, \smalltoken_2, \ldots, \smalltoken_{\numsmalltokens}\}$, or the \textit{large-scale} category, denoted by $\largetokens = \{\largetoken_1, \largetoken_2, \ldots, \largetoken_{\numlargetokens}\}$.

Small-scale tokens are pruned and replaced by their high-resolution counterparts. High-resolution tokens are generated by applying a shallow transformer encoder with $\numhrenc$ layers to the high-resolution image $\image$, where $\numhrenc = \numlrenc$ to ensure feature alignment. The small-scale tokens are then replaced by their corresponding high-resolution tokens (\cref{fig:pipeline}), resulting in an expanded set of high-resolution tokens, denoted as $\hrtokens = \{\hrtoken_1, \hrtoken_2, \ldots, \hrtoken_{\numhrtokens}\}$, where $\numhrtokens = 4\numsmalltokens$.

Although background can provide valuable contextual information, their feature encoding can be further optimized. To this end, we distill the background tokens $\bgtokens$ by spatially pooling every four neighboring tokens, resulting in pooled background tokens $\bgtokenspooled = \{\bgtokenpooled_1, \bgtokenpooled_2, \ldots, \bgtokenpooled_{\numbgtokenspooled}\}$. To avoid irregular pooling, some low-resolution tokens that cannot be grouped remain unchanged. As a result, $\numbgtokenspooled$ does not necessarily equal $\numbgtokens / 4$. By retaining these tokens, we better preserve their positional information, which facilitates subsequent positional encoding operations \cite{liu2022dabdetr}.

The remaining tokens in $\lrtokens$ representing large-scale individuals, $\largetokens$, remain unchanged. Next, we integrate the different tokens, including pooled background tokens and high-resolution tokens from selected regions. This results in scale-adaptive tokens, denoted as $\mrtokens = \{\bgtokenspooled, \largetokens, \hrtokens\}$. Compared to the uniform low-resolution tokens, this approach allocates feature details more efficiently, preserving different levels of detail for different individuals and regions. 

These tokens, $\mrtokens$, are then processed by another Transformer encoder with $\nummrenc$ layers and further decoded in subsequent stages. \cref{fig:pipeline} (bottom) provides an overview of the scale-adaptive encoder.

\vspace{0.4em}
\noindent\textbf{Decoder and prediction heads.} The decoder and prediction heads follow the architecture of our baseline approach as mentioned in \cref{subsec:overview}. For detailed architectures, please refer to the supplementary material (\cref{sec:supp_arch}).

\subsection{Training Losses}
\label{subsec:loss}
Similar to DETR \cite{carion2020end}, we first match our human predictions to \ac{gt} using Hungarian Matching, and we leverage bounding boxes, confidence scores, and projected joints during the matching. This matching step is necessary before computing the overall loss $\loss$. More details on the matching process are provided in \cref{sec:supp_arch}. The loss function, $\loss$, is a weighted sum of various terms, with each $\lambda$ representing a hyperparameter:
\vspace{-0.3em}
\begin{equation*}
\begin{aligned}
\loss = & \weightmap \lossmap + \weightdepth \lossdepth + \weightpose \losspose + \weightshape \lossshape \\
    &+ \weightjddd \lossjddd + \weightjdd \lossjdd + \weightbox \lossbox + \weightdet \lossdet.
\end{aligned}
\end{equation*}

\noindent\textbf{Scale map loss $\lossmap$.} To supervise the patch-level scale map, we apply focal loss \cite{Lin_2017_ICCV} to the predicted confidence $\conf$ and L1 loss to the predicted scale $s$. The total loss, $\lossmap$, is the sum of these two terms. The \ac{gt} for confidence and scale is precomputed, as described in \cref{subsec:scaleaware}.

\vspace{0.4em}
\noindent\textbf{Depth loss $\lossdepth$.} We supervise the root depth (\ie, the z-dimension) of the 3D joints $\joint$, regressed from the predicted mesh $\mesh$ (\cref{subsec:preliminaries}). Specifically, we normalize the depth based on the focal length, following \cite{facil2019cam}. The depth loss is calculated as $\lossdepth = \left|\frac{1}{\gtdepth} - \frac{\focal}{\gtfocal \depth}\right|$, where $\focal$ is the predefined focal length of our camera, and $\gtdepth$ and $\gtfocal$ are the \ac{gt} depth and focal length, respectively. See \cref{sec:supp_arch} for more details.

\begin{table*}[ht]
    \centering
    \caption{\textbf{Comparison with \ac{sota} methods on AGORA test set \cite{patel2021agora}.} ``Res.'' represents the input resolution. ``644$^*$'' means we use a base resolution of 644 and adaptively scale certain regions up to a maximum resolution of 1288.}
    \label{tab:sota_agora_test}
    \resizebox{.9\linewidth}{!}{ 
    \setlength{\tabcolsep}{1.4mm}
    \begin{tabular}{lccccccccccc} 
    \thickhline 
    Method & Res. & Time (ms) & MACs (G) & F1-Score $\uparrow$ & Precision $\uparrow$ & Recall $\uparrow$ & 
    MPJPE $\downarrow$ & MVE $\downarrow$ & NMJE $\downarrow$ & NMVE $\downarrow$ \\
    \hline
    ROMP \cite{sun2021monocular} & 512 & 38.7 & 43.6 & 0.91 & 0.95 & 0.88 & 108.1 & 103.4 & 118.8 & 113.6 \\
    BEV \cite{sun2022putting} & 512 & 50.6 & 48.9 & \cellcolor{third}0.93 & \cellcolor{third}0.96 & \cellcolor{third}0.90 & 105.3 & 100.7 & 113.2 & 108.3 \\
    PSVT \cite{qiu2023psvt} & 512 & - & - & \cellcolor{third}0.93 & - & - & 97.7 & 94.1 & 105.1 & 101.2 \\
    AiOS \cite{sun2024aios} & 1333 & 405.2 & 314.5 & \cellcolor{second}0.94 & \cellcolor{best}0.98 & \cellcolor{third}0.90 & \cellcolor{best}63.9 & \cellcolor{best}57.5 & \cellcolor{best}68.0 & \cellcolor{best}61.2 \\
    Multi-HMR \cite{multihmr2024} & 896 & 97.4 & 2075.1 & \cellcolor{third}0.93 & - & - & 82.8 & 77.6 & 89.0 & 83.4 \\
    Multi-HMR \cite{multihmr2024} & 1288 & 231.7 & 6104.6 & \cellcolor{best}0.95 & \cellcolor{second}0.97 & \cellcolor{best}0.93 & \cellcolor{second}65.3 & \cellcolor{second}61.1 & \cellcolor{second}68.7 & \cellcolor{second}64.3 \\
    \textbf{Ours} & 644$^*$ & 42.0 & 133.1 & \cellcolor{best}0.95  & \cellcolor{best}0.98 & \cellcolor{second}0.91 & \cellcolor{third}67.9 & \cellcolor{third}63.3 & \cellcolor{third}71.5 & \cellcolor{third}66.6 \\
    \thickhline
    \end{tabular}}
\vspace{-0.8em}
\end{table*}

\vspace{0.4em}
\noindent\textbf{Other losses.} For the pose parameters loss $\losspose$, shape parameters loss $\lossshape$, 3D joints loss $\lossjddd$, and projected 2D joints loss $\lossjdd$, we use the L1 distance. For the bounding box loss $\lossbox$, we combine L1 loss with GIoU loss \cite{rezatofighi2019generalized}. The detection loss $\lossdet$ uses focal loss \cite{Lin_2017_ICCV} to supervise the confidence scores of human queries.

\section{Experiments}

\subsection{Datasets and metrics}
\label{subsec:dataset}
We train our model on multi-person datasets including AGORA \cite{patel2021agora}, BEDLAM \cite{black2023bedlam}, COCO \cite{lin2014microsoft}, Crowdpose \cite{li2019crowdpose}, and MPII \cite{andriluka20142d}, as well as the single-person dataset H3.6M \cite{h36m_pami}. AGORA and BEDLAM are synthetic datasets with high-quality 3D mesh \ac{gt}s. For other multi-person datasets, we use pseudo \ac{gt}s from NeuralAnnot \cite{moon2022neuralannot} and only supervise projected 2D joints. 

For comparison with \ac{sota} methods, we evaluate our method on AGORA \cite{patel2021agora}, MuPoTS \cite{mehta2018single}, 3DPW \cite{vonMarcard2018}, and CMU Panoptic \cite{joo2015panoptic} datasets. Following previous works \cite{sun2021monocular, sun2022putting, multihmr2024, sun2024aios}, we report \acf{mve} and \ac{mpjpe} before and after \ac{pa}. On AGORA, we also report F1-Score, precision, and recall for evaluating detection accuracy; \ac{nmve} and \ac{nmje} that considered regression accuracy with detection accuracy. All metrics above, except detection accuracy, are reported in millimeters (mm). On MuPoTS-3D \cite{mehta2018single}, we report \ac{pck} using a threshold of 15 cm. To evaluate computational costs, we report \ac{macs} and the average inference time (ms) following \cite{multihmr2024} on one RTX 3090 GPU.

\subsection{Implementation Details}
\label{subsec:implementation}
The input image is resized to maintain its original aspect ratio, with a base resolution of $\max(\lrimgh, \lrimgw) = 644$ and $\max(\imgh, \imgw) = 1288$ for high-resolution when needed. We denote this mixed-resolution setting as ``644$^*$''. The Transformer layers in our encoder are initialized using pretrained DINOv2 \cite{oquab2023dinov2} with a patch size of $\patchsize = 14$. We use ViT-B architecture as ViT-L significantly increases inference time (42.0 ms vs. 74.2 ms). The Transformer used in the encoder has $\numlrenc = \numhrenc = 3$, and $\nummrenc = 9$ layers, respectively. 
Our decoder consists of 6 Transformer layers and 50 human queries. 
Training is conducted on 8 RTX 3090 GPUs with a total batch size of 40 for 60 epochs with an initial learning rate of $4e-5$. Please refer to \cref{sec:supp_training} for more details.

\begin{table}[t]
    \centering
    \caption{\textbf{Comparison with \ac{sota} methods on 3DPW \cite{vonMarcard2018} and MuPoTS \cite{mehta2018single} test sets.} ``Res.'' represents the input resolution. ```Crop'' refers to cropping individuals for single-person mesh estimation, \ie, a multi-stage approach. ``644$^*$'' means a base resolution of 644, with adaptive scaling up to 1288.}
    \label{tab:sota_3dpw_mupots}
    \resizebox{\linewidth}{!}{ 
    \setlength{\tabcolsep}{1mm}
    \begin{tabular}{lccccc} 
    \thickhline 
    \multirow{2}{*}{Method} & \multirow{2}{*}{Res.} & \multicolumn{2}{c}{3DPW} & \multicolumn{2}{c}{MuPoTS (PCK)} \\
    \cmidrule(r){3-4} \cmidrule(r){5-6}
     & & PA-MPJPE $\downarrow$ & MVE $\downarrow$ & All $\uparrow$ & Matched $\uparrow$ \\
    \hline
    CRMH \cite{jiang2020coherent} & 832 & - & - & 69.1 & 72.2\\
    3DCrowdNet \cite{choi2022learning} & Crop & 51.5 & 98.3 & 72.7 & 73.3\\
    ROMP \cite{sun2021monocular} & 512 & 47.3 & 93.4 & 69.9 & 72.2 \\
    BEV \cite{sun2022putting} & 512 & 46.9 & 92.3 & 70.2 & 75.2 \\
    PSVT \cite{qiu2023psvt} & 512 & 45.7 & 84.9 & - & - \\
    Multi-HMR \cite{multihmr2024} & 896 & 41.7 & 75.9 & 85.0 & 89.3 \\
    \rowcolor{mygray}
    \textbf{Ours} & 644$^*$ & \textbf{41.6} & \textbf{73.7} & \textbf{89.0} & \textbf{90.1} \\
    \thickhline
    \end{tabular}
    }
\vspace{-0.8em}
\end{table}

\subsection{Comparison to the State-of-the-arts}
\label{subsec:sota}
\begin{figure*}[t]
  \centering
  \includegraphics[width=.86\linewidth]{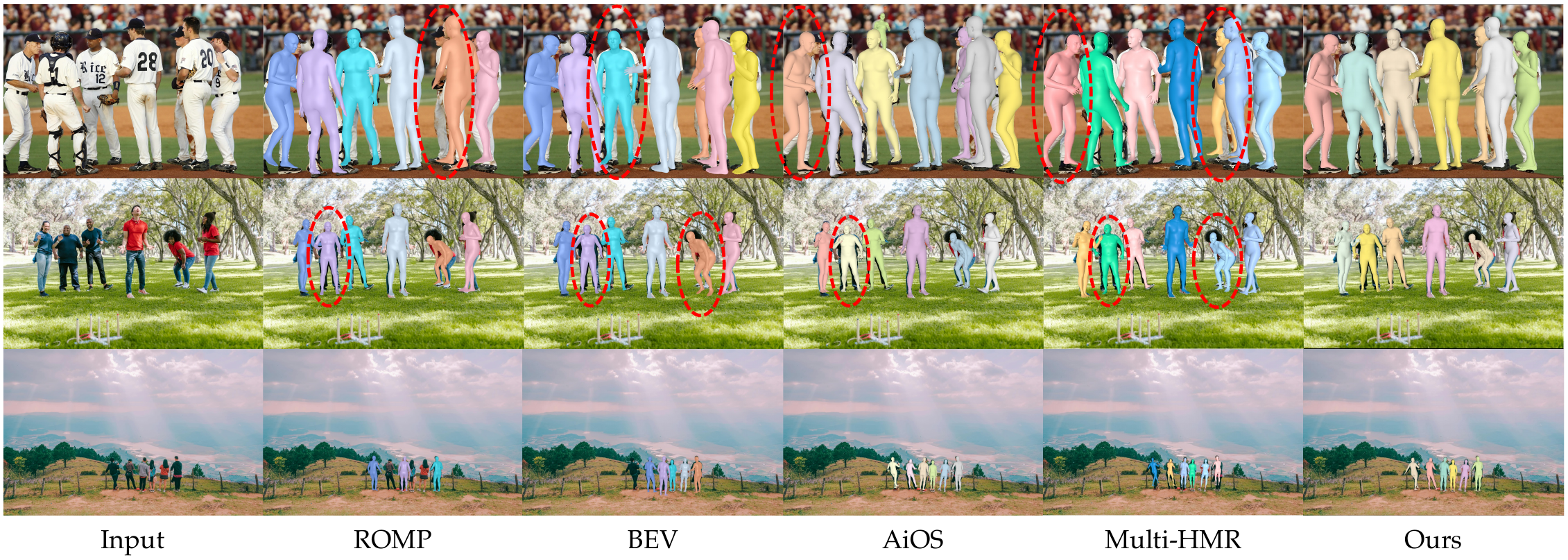}
  \caption{\textbf{Comparison with \ac{sota} methods \cite{sun2021monocular,sun2022putting,sun2024aios,multihmr2024} on in-the-wild images from the Internet.} Red dashed circles highlight areas with incorrect estimations. The third case is left blank due to the small scale of individuals. Please zoom in for details.}
  \label{fig:comparison}
  \vspace{-0.8em}
\end{figure*}

\noindent\textbf{Performance and Efficiency.} We conduct evaluations on the AGORA \cite{patel2021agora}, 3DPW \cite{vonMarcard2018}, MuPoTS \cite{mehta2017monocular}, and CMU Panoptic \cite{joo2015panoptic} datasets. \cref{tab:sota_agora_test} shows the results of our method compared to \ac{sota} methods on the AGORA \cite{patel2021agora} test set leaderboard. Our method, using a mixed resolution input with a base of 644, achieves performance on par with the top \ac{sota} methods \cite{multihmr2024,sun2024aios}, despite their use of higher resolutions and larger models (\eg, Multi-HMR use ViT-L \cite{oquab2023dinov2} model). These methods, however, result in significantly higher computational complexity, including much higher MACs and slower runtime, compromising real-time performance. Besides, these methods \cite{multihmr2024,sun2024aios} estimates SMPL-X \cite{pavlakos2019expressive}, from which SMPL \cite{loper2015smpl} mesh is derived. In contrast, our method maintains real-time computational efficiency while improving upon real-time methods ROMP \cite{sun2021monocular}, by nearly 40\%. This establishes our approach as \textbf{the leading real-time model}, setting a new standard by achieving \textbf{an unparalleled balance of performance and efficiency}.

\begin{table}[t!]
    \centering
    \caption{\textbf{Comparison with \ac{sota} methods on CMU Panoptic \cite{joo2015panoptic} test set.} We report \ac{mpjpe} for four activities and the average.}
    \label{tab:sota_cmu}
    \resizebox{\linewidth}{!}{ 
    \setlength{\tabcolsep}{0.6mm}
    \begin{tabular}{lccccc} 
    \thickhline 
    Method & Haggling $\downarrow$ & Mafia $\downarrow$ & Ultimatum $\downarrow$ & Pizza $\downarrow$ & Avg. $\downarrow$ \\
    \hline
    CRMH \cite{jiang2020coherent}  & 129.6 & 133.5 & 153.0 & 156.7 & 143.2\\
    3DCrowdNet \cite{choi2022learning} & 109.6 & 135.9 & 129.8 & 135.6 & 127.6 \\
    ROMP \cite{sun2021monocular} & 110.8 & 122.8 & 141.6 & 137.6 & 128.2 \\
    BEV \cite{sun2022putting} & 90.7 & 103.7  & 113.1 & 125.2 & 109.5 \\
    PSVT \cite{qiu2023psvt} & 88.7 & 97.9 & 115.2 & 121.2 & 105.7 \\
    \rowcolor{mygray}
    \textbf{Ours} & \textbf{67.9} & \textbf{78.5} & \textbf{95.8} & \textbf{94.6} & \textbf{84.2} \\
    \thickhline
    \end{tabular}}
\vspace{-0.8em}
\end{table}

\begin{figure}[t]
  \centering
  \includegraphics[width=.9\linewidth]{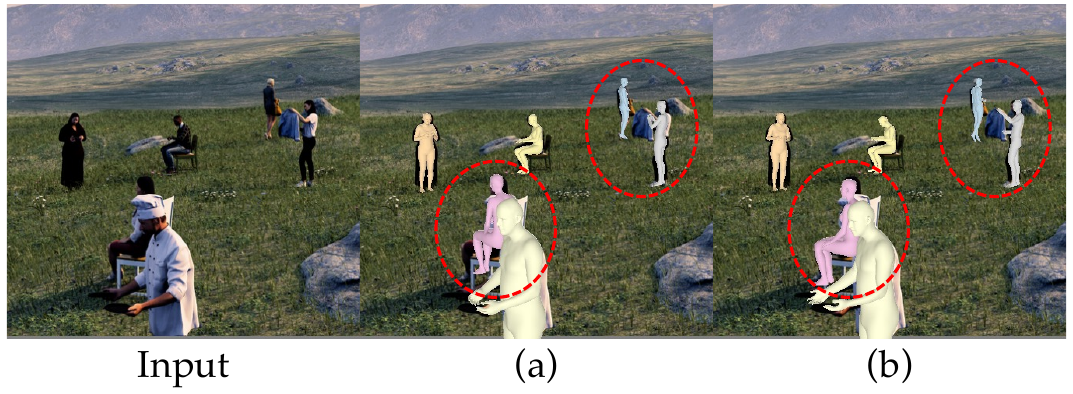}
  \caption{\textbf{Qualitative comparison of different resolutions for our baseline.} Resolution for the baseline: (a) 518, (b) 1288. Red dashed circles highlight differences; zoom in for details.}
  \label{fig:resolution}
\vspace{-1em}
\end{figure}

\cref{tab:sota_3dpw_mupots} presents the results of our method on the 3DPW dataset. Following Multi-HMR \cite{multihmr2024}, we finetune on the 3DPW training set. We further evaluate the generalization performance of our method on the MuPoTS and CMU Panoptic datasets, as shown in \cref{tab:sota_3dpw_mupots} and \cref{tab:sota_cmu}. For Panoptic dataset, we follow the common evaluation protocol \cite{sun2021monocular, sun2022putting, qiu2023psvt} and exclude Multi-HMR \cite{multihmr2024} due to potential inconsistencies in joint format. Our method consistently achieves the best results across all datasets, demonstrating exceptional generalization capability.

\vspace{0.4em}
\noindent\textbf{Qualitative Results.} \cref{fig:comparison} presents a visual comparison between our method and existing \ac{sota} approaches. The images are sourced from the Internet. Our method provides the most accurate overall estimations, especially in the third case, where individuals are far from the camera, resulting in very small scales. Despite this challenge, our method remains highly \textbf{robust}, delivering precise estimations. In contrast, ROMP \cite{sun2021monocular} and BEV \cite{sun2022putting} fail to detect all individuals, while AiOS \cite{sun2024aios} and Multi-HMR \cite{multihmr2024}, despite using higher-resolution models, show significant estimation errors.

\subsection{Ablation Study}
\label{subsec:ablation}
We conduct ablation studies using only the AGORA \cite{patel2021agora} and sampled BEDLAM \cite{black2023bedlam} datasets to validate model designs, and report results on their validation sets.

\begin{table}[t]
    \centering
    \caption{\textbf{Ablation studies on BEDLAM \cite{black2023bedlam} validation set.} We report \ac{mve} for different scale ranges and the average (Avg.).}
    \label{tab:ablation_bedlam}
    \resizebox{0.94\linewidth}{!}{ 
    \setlength{\tabcolsep}{0.8mm}
    \begin{tabular}{lcccccc} 
    \thickhline 
    Ablation & 0-20\% & 20-40\% & 40-60\% & 60-80\% & 80\%+ & Avg. \\
    \hline
    (a) Drop all & \cellcolor{best}59.9 & 55.1 & 61.5 & 66.6 & 70.8 & 57.2\\
    (b) No pooling & 60.3 & \cellcolor{second}54.5 & \cellcolor{best}57.0 & \cellcolor{best}59.8 & \cellcolor{second}64.1 & \cellcolor{second}56.1\\
    (c) Pooling$\times$2 & 60.7 & \cellcolor{second}54.5 & 57.5 & 62.1 & 66.5 & 56.3\\
    (d) \textbf{Ours} & \cellcolor{second}60.0 & \cellcolor{best}54.4 & \cellcolor{second}57.2 & \cellcolor{second}60.3 & \cellcolor{best}62.7 & \cellcolor{best}56.0 \\
    \thickhline
    \end{tabular}}
\vspace{-1em}
\end{table}

\vspace{0.4em}
\noindent\textbf{Impact of resolution.} \cref{fig:ablation_scale_fps} shows baseline results with different input resolutions on the AGORA \cite{patel2021agora} validation set. As resolution increases, the error across individuals in different scale ranges consistently decreases, with the most significant drop (35mm) in the smallest scale range. In contrast, the improvement in the large-scale range is less pronounced. However, the performance gain comes at the cost of a significant drop in computational efficiency, reducing the frame rate from real-time (32 FPS) to 5 FPS. \cref{fig:resolution} shows an example, clearly illustrating that higher resolution greatly improves estimation for small-scale individuals.

\begin{figure*}[t]
  \centering
  \includegraphics[width=.88\linewidth]{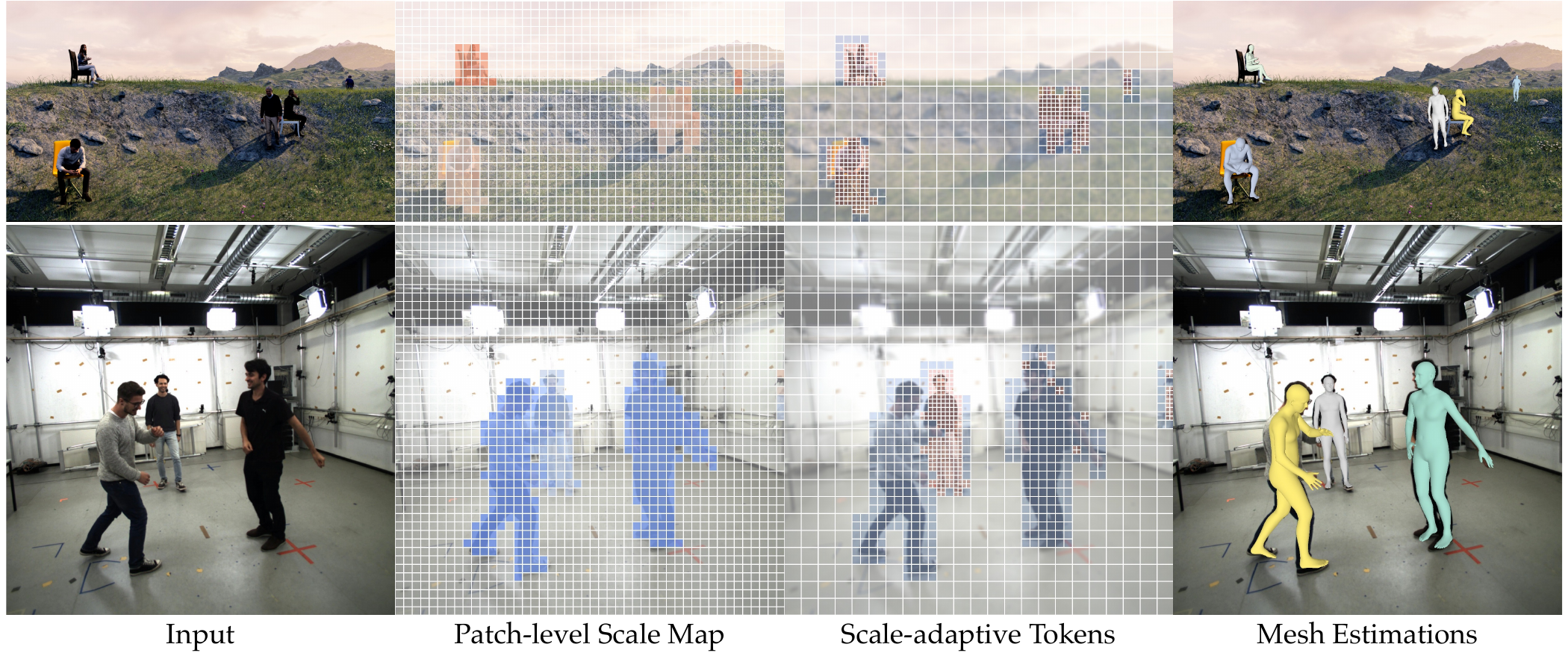}
  \caption{\textbf{Visualization of our predicted patch-level scale maps $\scalemap$ and scale-adaptive tokens $\mrtokens$.} We show the predicted scale maps, scale-adaptive tokens, and the estimated meshes overlaid on the image. In the scale map, colors represent scale values, with uncolored areas indicating background patches. A gradient from \textcolor{myblue}{blue} (large-scale individuals) to \textcolor{myrose}{rose} (small-scale individuals) illustrates the scale distribution. We visualize scale-adaptive tokens as patches of different sizes and colors on the image, with low-resolution patches blurred.}
  \label{fig:scalemap}
\vspace{-0.8em}
\end{figure*}

\begin{figure*}[t]
  \centering
  \includegraphics[width=.88\linewidth]{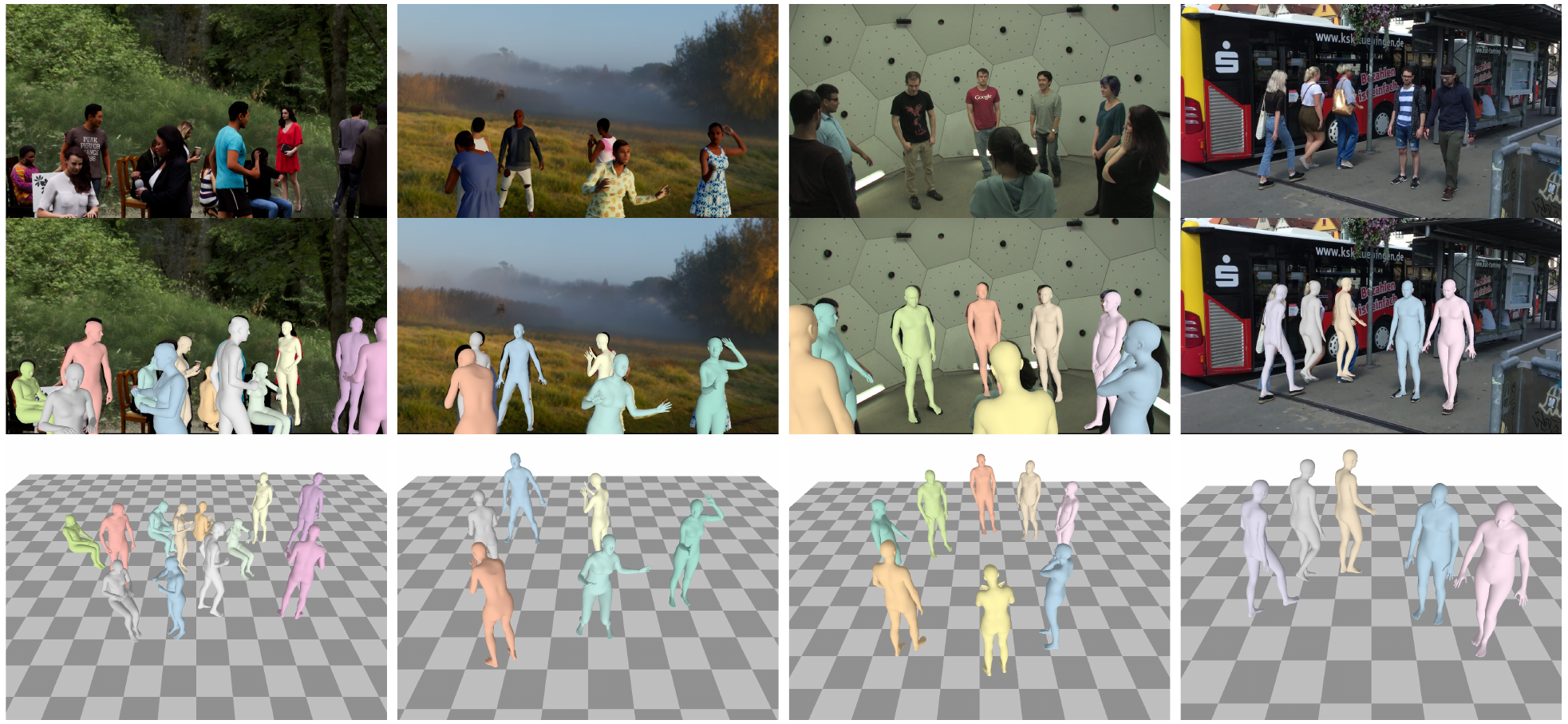}
  \caption{\textbf{Qualitative results of our method.} We visualize four cases from AGORA \cite{patel2021agora}, BEDLAM \cite{black2023bedlam}, CMU Panoptic \cite{joo2015panoptic} and 3DPW \cite{vonMarcard2018} datasets, displaying the input, the estimated mesh overlay, and an elevated view (top to bottom).}
  \label{fig:quality_results}
\vspace{-0.8em}
\end{figure*}

\vspace{0.4em}
\noindent\textbf{Background tokens} $\bgtokens$\textbf{.} We investigate the impact of background tokens on model performance and report the \ac{mve} across different scale ranges on the BEDLAM validation set in \cref{tab:ablation_bedlam}. In (a), we discard all background tokens and observe stable performance at the smallest scale, but a decline as the scale increases, indicating that background context becomes more important for larger scales. In (b), skipping pooling for background tokens yields nearly the same performance but introduces higher computational cost. In (c), applying pooling twice with the same grouping strategy results in a performance drop for large-scale instances, highlighting the significance of background context for estimation. We hypothesize that large-scale people face more truncation, relying more on global context which requires further exploration. We provide additional efficiency comparisons in \cref{sec:supp_exp}, showing that pooling once achieves the best balance between performance and efficiency.

\subsection{Qualitative Results}
\label{subsec:quality}
\cref{fig:scalemap} visualize the predicted patch-level scale map and scale-adaptive tokens by our model, enabling dynamic adjustment of tokens for accurate estimations. \cref{fig:quality_results} presents additional visualizations of predicted meshes, demonstrating the effectiveness of our method in various scenarios. Please refer to \cref{sec:supp_exp} for more qualitative results and discussion.

\section{Conclusion}
\vspace{-0.1em}
We present a novel one-stage framework for real-time multi-person 3D mesh estimation from an RGB image. By introducing scale-adaptive tokens that dynamically adjust based on the relative scale of individuals, our method effectively balances efficiency and accuracy, retaining the benefits of high-resolution processing while achieving real-time performance. Experiments demonstrate that our method is the \textbf{best real-time model}, offering competitive performance and superior generalization compared to \ac{sota}s. Our \textbf{limitations} include the lack of height-aware scale estimation, which may cause depth errors, and the current focus on body-only mesh estimation, extendable to whole-body in the future.

\clearpage

\section*{Acknowledgments}
This work was supported by 2022ZD0114904 and NSFC-6247070125. We thank Yiyang Wang for the valuable discussions during the early stage of this work. 

{
    \small
    \bibliographystyle{ieeenat_fullname}
    \bibliography{ref}
}

\clearpage
\setcounter{page}{1}
\maketitlesupplementary
\renewcommand\thesection{\Alph{section}}  
\renewcommand\thefigure{\Alph{section}\arabic{figure}} 
\renewcommand\thetable{\Alph{section}\arabic{table}}
\renewcommand\theequation{\Alph{section}\arabic{equation}}
\setcounter{section}{0}
\setcounter{figure}{0}
\setcounter{table}{0}
\setcounter{equation}{0}

In \cref{sec:supp_arch}, we elaborate on the implementation details of our proposed method and the experimental setups. We provide additional results and extended discussions in \cref{sec:supp_exp}.

\section{Implementation Details}
\subsection{Model Architecture} \label{sec:supp_arch}

\begin{figure*}[t]
  \centering
  \includegraphics[width=.8\linewidth]{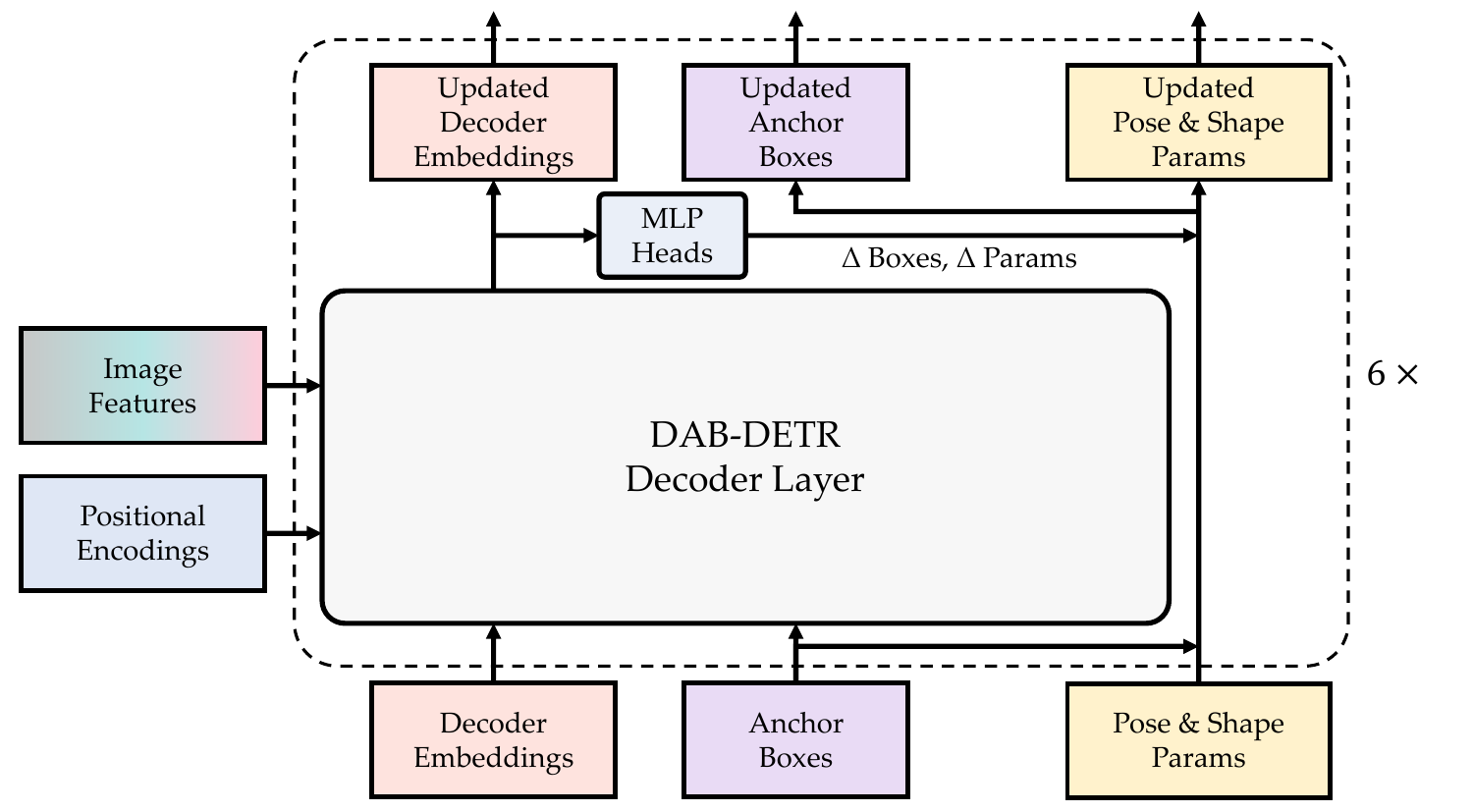}
  \caption{\textbf{Illustration of our decoder architecture.} Queries consist of \textit{decoder embeddings} and \textit{anchor boxes} following DAB-DETR \cite{liu2022dabdetr}. Besides updating anchor boxes, we also update SMPL parameters using corresponding prediction heads.}
  \label{fig:supp_decoder}
\vspace{-0.6em}
\end{figure*}

\begin{table*}[ht]
    \centering
    \caption{\textbf{Speed-accuracy trade-off among different ablation settings.} We conduct our studies on BEDLAM \cite{black2023bedlam} validation set, reporting the average number of different tokens, inference runtime and \ac{mve}.}
    \label{tab:supp_ablation_speed_accuracy}
    \setlength{\tabcolsep}{1.2mm}{
    \begin{tabular}{lccccc} 
    \thickhline 
    \multirow{2}{*}{Ablation} & \multicolumn{3}{c}{Number of tokens} & \multirow{2}{*}{Runtime (ms)} & \multirow{2}{*}{MVE $\downarrow$}\\
    \cline{2-4}
    & High-resolution & Low-resolution & Background \\
    \hline
    \multicolumn{6}{c}{Single Resolution}\\
    \hline
    (a) Res. 1288 & \multicolumn{3}{c}{4784} & 174.9 & 53.2\\
    (b) Res. 644 & \multicolumn{3}{c}{1196} & 42.3 & 63.3\\
    \hline
    \multicolumn{6}{c}{Background Tokens $\bgtokens$}\\
    \hline
    (c) Drop all & 493 & 40 & 0 & 39.7 & 57.2\\
    (d) No pooling & 493 & \multicolumn{2}{c}{1073} & 54.1 & 56.1\\
    (e) Pooling$\times$2 & 493 & 94 & 100 & 41.6 & 56.3\\
    \hline
    \rowcolor{mygray}
    (f) \textbf{Ours} & 493 & 94 & 245 & 42.0 & 56.0\\
    \thickhline
    \end{tabular}}
\end{table*}

\noindent\textbf{Encoder-decoder design.} Our model follows an encoder-decoder design based on previous works \cite{liu2022dabdetr, multihmr2024}, with the vanilla encoder replaced by our proposed scale-adaptive encoder. For the decoder, following \cite{liu2022dabdetr}, human queries consist of two learnable components: the \textit{content} part and the \textit{positional} part, also known as \textit{decoder embeddings} and \textit{anchor boxes} \cite{liu2022dabdetr}. Following \cite{liu2022dabdetr}, anchor boxes are refined layer-by-layer by predicting residual values via a prediction head. Additionally, we initialize mean SMPL pose and shape parameters used in \cite{kolotouros2019learning, goel2023humans, multihmr2024} and update them with a similar procedure, as illustrated in \cref{fig:supp_decoder}. Predictions of 3D translation $\transl$ are regressed from updated decoder embeddings without iteratively updating, which is not included in \cref{fig:supp_decoder}. Finally, all human predictions are matched to \ac{gt}s before computing training losses. We adopt the Hungarian algorithm following previous works \cite{carion2020end, sun2024aios}. The matching cost is computed as a weighted sum of $\lossbox$, $\lossdet$, and $\lossjdd$, with the weights sharing the same values as those in \cref{sec:supp_training}.

\vspace{0.8em}
\noindent\textbf{Camera model.} To leverage 2D annotations for supervision, we adopt a pinhole camera model to project 3D joints onto the image plane. Given the focal length $\focal$ and principal point $(\principalu, \principalv)$, a 3D point $(x, y, z)$ is projected to the image coordinates $(u, v)$ as follows:

\begin{equation} u = \frac{\focal \times x}{z} + \principalu, \quad v = \frac{\focal \times y}{z} + \principalv. \end{equation}
Following \cite{sun2022putting, multihmr2024}, we assume a standard camera with a fixed \ac{fov} of $60^\circ$. Given $\imgsize$ as the longer side of the image, the focal length is predefined as $\focal = \imgsize / (2\tan{(FOV/2)})$. The principal point $(\principalu, \principalv)$ is located at the center of the image.

\subsection{Training} \label{sec:supp_training}

The confidence and scale thresholds corresponding to the scale map are set to $\mapthreshconf = 0.3$ and $\mapthreshscale = 0.5$, respectively. 
The loss weights are set to $\weightmap = 4$, $\weightdepth = 0.5$, $\weightpose = 5$, $\weightshape = 3$, $\weightjddd = 8$, $\weightjdd = 40$, $\weightbox = 2$ and $\weightdet = 4$. 
We train our model with AdamW \cite{Loshchilov2017DecoupledWD}, with weight decay set to $1e-4$. The initial learning rate for the pretrained parameters is set to $2e-5$, while for other parameters, it is set to $4e-5$. The model is trained for 60 epochs with a total batch size of 40, which takes around a week on 8 RTX 3090 GPUs.

\subsection{Datasets} \label{sec:supp_datasets}
We briefly introduce the datasets used for training or evaluation.

\vspace{0.8em}
\noindent\textbf{AGORA} \cite{patel2021agora} is a synthetic dataset known for its high realism and diverse scenarios. Due to its highly accurate \ac{gt}s annotated in both SMPL \cite{loper2015smpl} and SMPL-X \cite{pavlakos2019expressive}, AGORA has become an essential benchmark for evaluating 3D human mesh estimation models. It contains approximately 14K images with 107K instances for training, 1K images with 8K instances for validation, and 3K images for testing.

\vspace{0.8em}
\noindent\textbf{BEDLAM} \cite{black2023bedlam} is a large-scale, synthetic video dataset that includes a diversity of body shapes, motions, skin tones, hair, and clothing. 
The dataset contains approximately 286K images with 951K instances for training and 29K images with 96K instances for validation. For our ablation study, we uniformly downsample the training set by a factor of 6, resulting in 48K images with 159K instances. We do not use the test set because the SMPL format is not currently supported by the leaderboard.

\vspace{0.8em}
\noindent\textbf{COCO} \cite{lin2014microsoft}, \textbf{Crowdpose} \cite{li2019crowdpose}, and \textbf{MPII} \cite{andriluka20142d} are real-world multi-person datasets widely used for 2D human pose estimation tasks. We use these datasets for training to enhance the generalization capability of our model on real-world images by using pseudo annotations from NeuralAnnot \cite{moon2022neuralannot} and only supervise projected 2D joints due to 3D ambiguity and their label noisiness. We uniformly downsample COCO by a factor of 4, resulting in 16K images with 66K instances for training. For Crowdpose, we use 10K images with 36K instances, and for MPII, we use 17K images with 29K instances.

\vspace{0.8em}
\noindent\textbf{H3.6M} \cite{h36m_pami} is an indoor single-person dataset with 3D pose annotations. It contains videos of common activities performed by professional actors. We uniformly downsample its training set by a factor of 10 and use 31K images.

\vspace{0.8em}
\noindent\textbf{3DPW} \cite{vonMarcard2018} is an in-the-wild dataset with 3D mesh annotations. It contains approximately 17K images for training and 24K images for testing. Following \cite{sun2021monocular, sun2022putting, multihmr2024}, we use the training set to finetune our model before evaluating the test set. 

\vspace{0.8em}
\noindent\textbf{MuPoTS} \cite{mehta2018single} is a real-world multi-person 3D pose dataset composed of more than 8K frames from 20 scenes, each containing up to three subjects, annotated with 3D pose. Following previous works \cite{sun2022putting, multihmr2024}, we only use it to evaluate the generalization capability of our model.

\vspace{0.8em}
\noindent\textbf{CMU Panoptic} \cite{joo2015panoptic} is an indoor multi-person dataset providing 3D pose annotations. It contains 4 sequences of multiple people engaging in different social activities, with approximately 9K images. Following previous works \cite{sun2021monocular, sun2022putting, multihmr2024}, we only use it to evaluate the generalization capability of our model.

\section{Extended Results} \label{sec:supp_exp}

\subsection{Ablation Study}
\label{supp:subsec:ablation}
\noindent\textbf{Speed-accuracy trade-off.} We evaluate the speed-accuracy trade-off across various ablation models. Specifically, we compare single-resolution baselines with our model, which adopts different processing strategies for background tokens ($\bgtokens$). In \cref{tab:supp_ablation_speed_accuracy}, we report the average number of tokens, inference runtime, and the \ac{mve} metric on BEDLAM \cite{black2023bedlam} validation set. The average number of tokens is included to highlight the impact of image tokens on inference speed.

In ablation (a), our baseline model with a resolution of 1288 achieves the lowest estimation error but suffers from redundant image tokens, leading to extremely slow inference. In contrast, ablation (b) shows faster inference but with poor performance, \ie much higher \ac{mve}. In (d), simply replacing small-scale tokens with their high-resolution counterparts brings a noticeable boost in performance with additional overhead, where some low-resolution tokens are still redundant. In (f), our proposed method of pooling background tokens $\bgtokens$ once counteracts the overhead brought by high-resolution tokens, yielding similar performance. However, further reducing $\bgtokens$ brings no significant acceleration and may result in a potential performance drop, as illustrated in (c) and (e). These results demonstrate that our scale-adaptive strategy achieves the best speed-accuracy trade-off, making our method the \textbf{best real-time model}.

\begin{table}[t]
    \centering
    \caption{\textbf{Effect of resolution.} We study the impact of resolution using both single-resolution (baseline) and mixed-resolution (our method, with ``*'') settings on AGORA \cite{patel2021agora} and BEDLAM \cite{black2023bedlam}, reporting \ac{mve} for different scale ranges and the average (Avg.).}
    \label{tab:supp_ablation_resolution}
    \resizebox{\linewidth}{!}{ 
    \setlength{\tabcolsep}{2mm}
    \begin{tabular}{lcccccc}
    \thickhline
     & Res. & 0-30\% & 30-50\% & 50-70\% & 70\%+ & Avg. \\
    \hline
    \multirow{6}{*}{\rotatebox{90}{BEDLAM}}  
        & 644 & 65.1 & 61.0 & 58.4 & 64.6 & 63.3 \\
        & 896 & 57.8 & 54.7 & 54.7 & 59.3 & 56.5 \\
        & 1288 & 55.2 & 50.4 & 51.6 & 56.0 & 53.2 \\
        & 448* & 60.3 & 60.3 & 61.0 & 68.6 & 60.5 \\
        & 644* & 55.6 & 56.0 & 57.6 & 63.1 & 56.0 \\
        & 896* & 51.1 & 51.3 & 53.7 & 58.6 & 51.4 \\
    \hline
     & Res. & 0-10\% & 10-20\% & 20-30\% & 30\%+ & Avg. \\
    \hline
    \multirow{6}{*}{\rotatebox{90}{AGORA}}  
        & 644 & 100.8 & 77.2 & 59.2 & 53.0 & 72.0 \\
        & 896 & 91.4 & 71.3 & 55.3 & 50.1 & 67.0 \\
        & 1288 & 82.2 & 64.9 & 52.4 & 48.2 & 61.9 \\
        & 448* & 94.6 & 74.1 & 58.5 & 54.7 & 70.0 \\
        & 644* & 84.6 & 68.5 & 57.3 & 52.7 & 65.5 \\
        & 896* & 76.5 & 63.7 & 54.1 & 49.9 & 61.0 \\ 
    \thickhline
    \end{tabular}}
\end{table}

\begin{table}[t]
    \centering
    \caption{\textbf{Dataset comparisons.} We report the number of instances for different scale ranges on different datasets, as well as \ac{mve} comparisons between our method (with ``*'') and the baseline.}
    \label{tab:supp_ablation_datasets}
    \resizebox{\linewidth}{!}{ 
    \setlength{\tabcolsep}{2mm}
    \begin{tabular}{lcccccc}
    \thickhline
     &  & 0-30\% & 30-50\% & 50-70\% & 70\%+ & Avg. \\
    \hline
    \multirow{3}{*}{\rotatebox{90}{BEDLAM}}  
        & Count & 54041 & 36362 & 4586 & 1497 & - \\
        & 644* & 55.6 & 56.0 & 57.6 & 63.1 & 56.0 \\
        & 644 & 65.1 & 61.0 & 58.4 & 64.6 & 63.3 \\
    \hline
    \multirow{3}{*}{\rotatebox{90}{3DPW}}  
        & Count & 5121 & 10113 & 12299 & 7982 & - \\
        & 644* & 86.2 & 74.2 & 70.5 & 70.2 & 73.7 \\
        & 644 & 93.6 & 77.8 & 68.7 & 68.3 & 74.8 \\
    \hline
     &  & 0-10\% & 10-20\% & 20-30\% & 30\%+ & Avg. \\
    \hline
    \multirow{3}{*}{\rotatebox{90}{AGORA}}  
        & Count & 1556 & 2976 & 1986 & 1274 & - \\
        & 644* & 84.6 & 68.5 & 57.3 & 52.7 & 65.5 \\
        & 644 & 100.8 & 77.2 & 59.2 & 53.0 & 72.0\\
    \hline
     &  & \multicolumn{2}{c}{0-40\%} & \multicolumn{2}{c}{40\%+} & Avg. \\
    \hline
    \multirow{3}{*}{\rotatebox{90}{Panoptic}}  
        & Count & \multicolumn{2}{c}{31592} & \multicolumn{2}{c}{11172} & - \\
        & 644* & \multicolumn{2}{c}{85.2} & \multicolumn{2}{c}{81.0} & 84.2 \\
        & 644 & \multicolumn{2}{c}{90.0} & \multicolumn{2}{c}{82.6} & 88.2\\
    \thickhline
    \end{tabular}}
\end{table}

\begin{table}[t]
    \centering
    \caption{\textbf{Comparison of different scale thresholds on BEDLAM \cite{black2023bedlam} validation set.} We report \ac{mve} for different scale ranges, average (Avg.) \ac{mve} and inference runtime (ms).}
    \label{tab:supp_ablation_bedlam}
    \resizebox{\linewidth}{!}{ 
    \setlength{\tabcolsep}{1.5mm}
    \begin{tabular}{lcccccc} 
    \thickhline
    \multirow{2}{*}{$\mapthreshscale$} & \multicolumn{5}{c}{MVE $\downarrow$} & \multirow{2}{*}{Time (ms)} \\
    \cline{2-6}
     & 0-30\% & 30-50\% & 50-70\% & 70\%+ & Avg. & \\
    \hline
    (a) 0.0 & 67.0 & 60.4 & 58.6 & 64.0 & 64.0 & 37.2\\
    (b) 0.3 & 60.0 & 61.9 & 59.9 & 65.8 & 60.8 & 40.9\\
    \rowcolor{mygray}
    (c) \textbf{0.5} & & & & & & \\
    \rowcolor{mygray}
    \textbf{(Ours)} & \multirow{-2}{*}{55.6} & \multirow{-2}{*}{56.0} & \multirow{-2}{*}{57.6} & \multirow{-2}{*}{63.1} & \multirow{-2}{*}{56.0} & \multirow{-2}{*}{42.0}\\
    (d) 0.7 & 55.9 & 56.1 & 58.2 & 65.2 & 56.2 & 44.1\\
    (e) 1.0 & 58.5 & 56.8 & 58.5 & 63.9 & 57.9 & 46.4\\
    \thickhline
    \end{tabular}}
\vspace{-0.8em}
\end{table}

\begin{figure*}[ht]
  \centering
  \includegraphics[width=\linewidth]{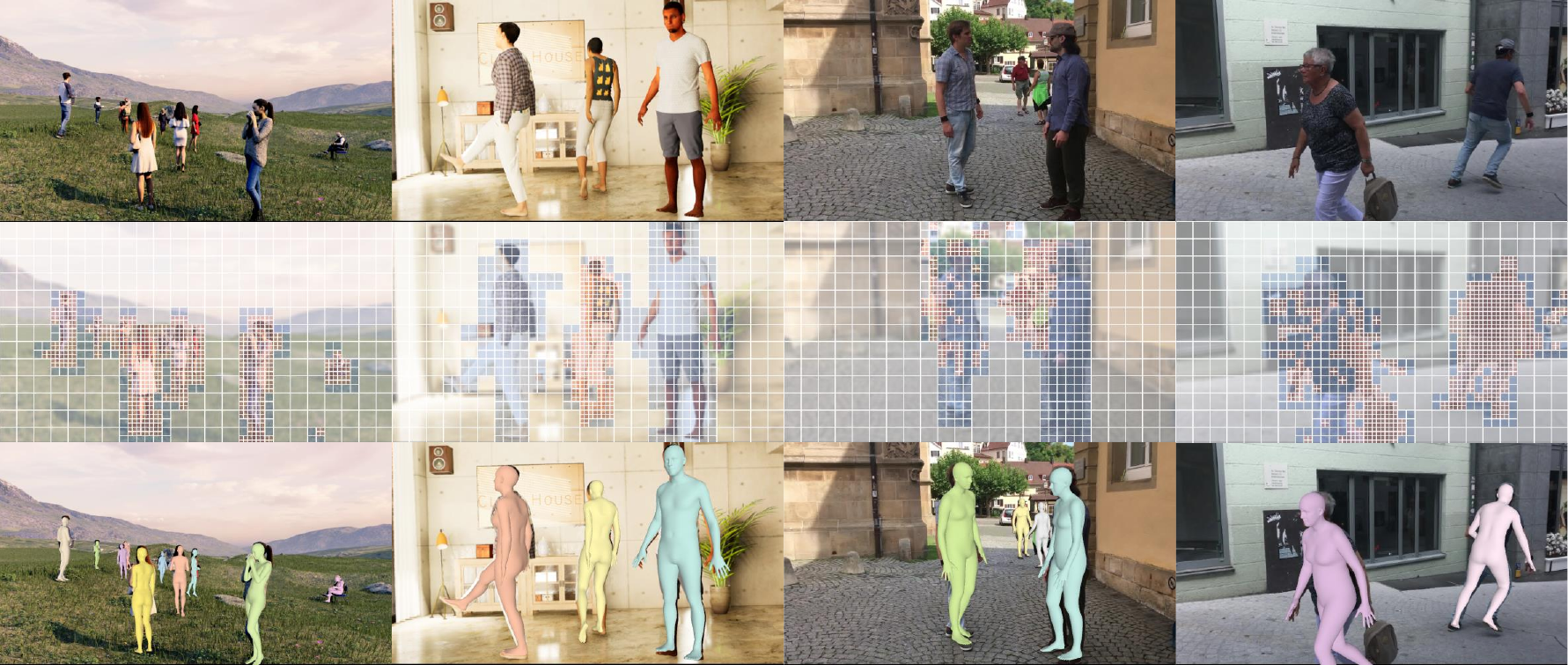}
  \caption{\textbf{Additional visualization of scale-adaptive tokens $\mrtokens$.} We display the input, scale-adaptive tokens and the estimated mesh overlay (top to bottom). Tokens are visualized the same way as \cref{fig:scalemap}. The first two columns are qualitative results on synthetic datasets \cite{patel2021agora, black2023bedlam}, and the last two columns are on 3DPW \cite{vonMarcard2018}.}
  \label{fig:supp_sat}
\vspace{-1em}
\end{figure*}

\vspace{0.8em}
\noindent\textbf{Effect of resolution.} To study the impact of resolution, we train our single-resolution baseline and our proposed method with scale-adaptive tokens using different resolution settings, reporting estimation errors (MVE) on AGORA \cite{patel2021agora} validation set and BEDLAM \cite{black2023bedlam} validation set in \cref{tab:supp_ablation_resolution}. As input resolution increases, both single-resolution (baseline) and mixed-resolution (our method, with ``*'') settings show accuracy improvements across individuals in different scale ranges. Compared to the corresponding baseline, our method greatly reduces the estimation error in small-scale instances. These results further demonstrate the importance of higher resolution, as it leads to better outcomes, and highlight the effectiveness of our mixed-resolution strategy.

\vspace{0.8em}
\noindent\textbf{Scale distribution.} We report the number of instances for each scale range on AGORA \cite{patel2021agora} validation set, BEDLAM \cite{black2023bedlam} validation set, 3DPW \cite{vonMarcard2018} test set and CMU Panoptic \cite{joo2015panoptic} test set in \cref{tab:supp_ablation_datasets} (MuPoTS \cite{mehta2018single} is not included due to the lack of bounding box annotations). To study the impact of scale-adaptive tokens, we evaluate our method (644*, mixed resolution) and the corresponding single-resolution baseline (644) on these datasets. As shown, AGORA \cite{patel2021agora} and BEDLAM \cite{black2023bedlam} contain more small-scale instances, where our method outperforms baseline with reduced \ac{mve}. This improvement is also seen on small-scale instances in 3DPW \cite{vonMarcard2018} and CMU Panoptic \cite{joo2015panoptic}. However, 3DPW's larger-scale instances show no improvement, likely due to the model's limited capability.

\vspace{0.8em}
\noindent\textbf{Scale threshold} $\mapthreshscale$\textbf{.} To further study the impact of high-resolution tokens, we conduct experiments on various scale thresholds $\mapthreshscale$ while retaining the pooling of background tokens. $\mapthreshscale = 0$ denotes that no high-resolution tokens are used. Results are shown in \cref{tab:supp_ablation_bedlam}. Compared to (a), our method (c) achieves a consistent error reduction across different scale ranges, indicating that introducing sufficient high-resolution tokens eases the estimation challenge on small-scale instances and also allows the model to deal better with large-scale instances. In (b), although the improvement in the scale range of 0-30\% is significant, the decrease in high-resolution samples increases learning difficulty during training and potentially leads to worse performance on larger-scale instances than (a). (d) and (e) indicate that too large $\mapthreshscale$ decreases efficiency with longer interence time cost without improving accuracy. In (d), a large scale threshold ignores plenty of background context since the high-resolution tokens are encoded independently for $\numhrenc$ layers, leading to a performance decline, which is consistent with our findings in \cref{subsec:ablation}. In (e), when all humans are processed with high-resolution tokens (\ie $\mapthreshscale = 1.0$), training becomes unstable and results are worse. In general, $\mapthreshscale = 0.5$ achieves the best speed-accuracy trade-off.

\vspace{0.8em}
\noindent\textbf{Accuracy and effect of scale map prediction.} We evaluate our scale map prediction in \cref{tab:supp_ablation_scale_map} which achieves high prediction accuracy, with 0.98 F1-Score. To further analyze its impact on the final mesh estimation, we replace the predicted scale map with \ac{gt} scale map during inference time. The average \ac{mve} slightly improves, indicating our scale predictions are accurate with minimal impact on accuracy. For scales over 50\%, we find that the predicted scale map outperforms \ac{gt} due to underestimating scales in some cases, assigning more instances of high-resolution tokens, thus improving results.

\begin{table}[t]
    \centering
    \caption{\textbf{Accuracy and effect of scale prediction on BEDLAM \cite{black2023bedlam} validation set.} We report F1-Score (F1) and \ac{mae} for evaluating scale map accuracy. To analyze its impact on mesh estimation, we replace the predicted scale map with \ac{gt} and report \ac{mve} for different scale ranges and the average (Avg.).}
    \label{tab:supp_ablation_scale_map}
    \resizebox{\linewidth}{!}{ 
    \setlength{\tabcolsep}{1.5mm}
    \begin{tabular}{lccccccc} 
    \thickhline
    Scale& \multirow{2}{*}{F1 $\uparrow$} & \multirow{2}{*}{MAE $\downarrow$} & \multicolumn{5}{c}{MVE $\downarrow$}\\
    \cline{4-8}
    map & & & 0-30\% & 30-50\% & 50-70\% & 70\%+ & Avg. \\
    \hline
    Pred. & 0.98 & 0.056 & 55.6 & 56.0 & 57.6 & 63.1 & 56.0 \\
    GT & - & - & 55.4 & 55.8 & 58.6 & 63.2 & 55.8 \\
    \thickhline
    \end{tabular}}
\vspace{-0.8em}
\end{table}

\begin{figure*}[t]
  \centering
  \includegraphics[width=\linewidth]{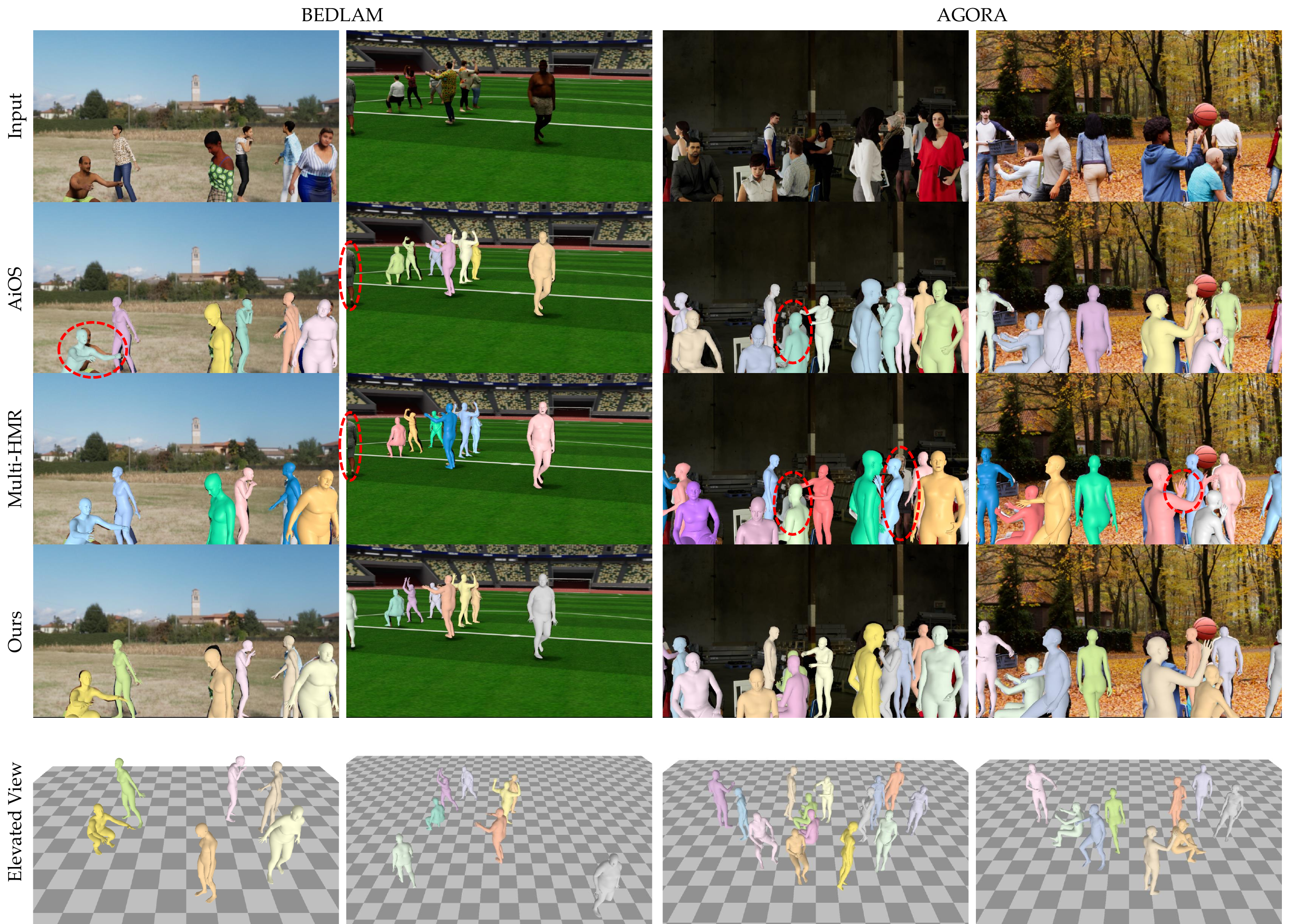}
  \caption{\textbf{Comparison on synthetic images.} We compare our method with other \ac{sota} methods \cite{sun2024aios,multihmr2024} on BEDLAM \cite{black2023bedlam} (left) and AGORA \cite{patel2021agora} (right). Red dashed circles highlight areas with 2D misalignment or misdetection. The last row shows the elevated view of our estimations. Please zoom in for details.}
  \label{fig:supp_synthetic}
\vspace{-1em}
\end{figure*}

\begin{figure*}[t]
  \centering
  \includegraphics[width=\linewidth]{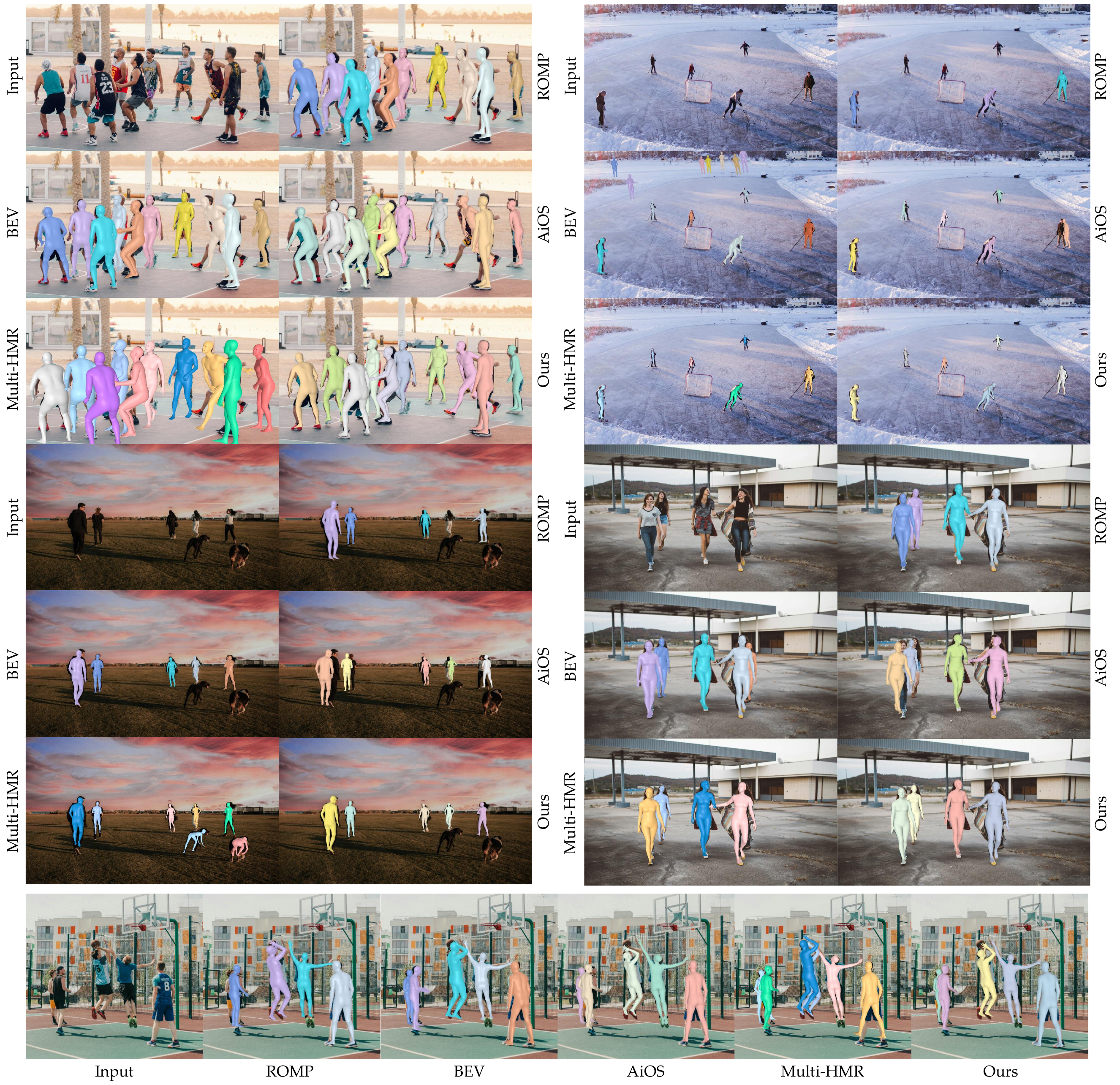}
  \caption{\textbf{Comparison on real-world images.} We compare our method with \ac{sota} methods \cite{sun2021monocular,sun2022putting,sun2024aios,multihmr2024} on in-the-wild images from the Internet. Our method outperforms all of them, especially in small-scale cases. Please zoom in for details.}
  \label{fig:supp_internet}
\vspace{-1em}
\end{figure*}

\begin{figure*}[t]
  \centering
  \includegraphics[width=\linewidth]{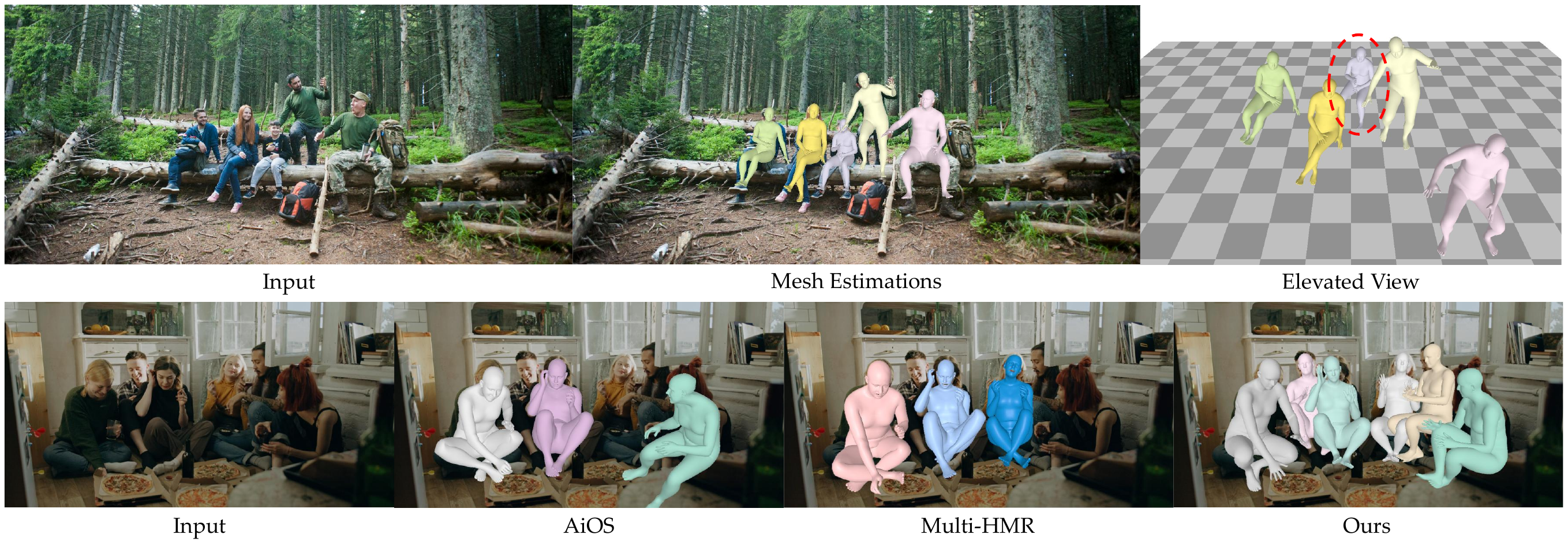}
  \caption{\textbf{Failure cases.} The top row shows an example of improper depth reasoning for the child. The bottom row shows poor estimation results of current \ac{sota} methods in complex human poses and scenarios.}
  \label{fig:supp_failure}
\vspace{-1em}
\end{figure*}

\subsection{Additional Qualitative Results}
\label{supp:subsec:quality}

\noindent\textbf{Scale-adaptive tokens.} We present additional visualized examples of our scale-adaptive tokens $\mrtokens$ in \cref{fig:supp_sat}. The last two cases include instances with scales near the scale threshold $\mapthreshscale$, resulting in mixed-resolution token representations for those individuals. Nevertheless, our model still produces satisfactory predictions, demonstrating the robustness and consistency of the features learned across different resolution levels.

\vspace{0.8em}
\noindent\textbf{\ac{sota} comparisons.} \cref{fig:supp_synthetic} and \cref{fig:supp_internet} present visual comparisons between our method and existing \ac{sota} approaches \cite{sun2021monocular,sun2022putting,sun2024aios,multihmr2024} on synthetic images and real-world images, respectively. Our method demonstrates a strong generalization capability with accurate estimations across different scenarios. Specifically, our method can accurately estimate individuals across different scales, whereas other methods may fail to detect very small individuals or produce inaccurate estimations. See \cref{fig:supp_internet} for qualitative examples illustrating this advantage.

\vspace{0.8em}
\noindent\textbf{Failure cases.} \cref{fig:supp_failure} (top) indicates that our method can result in unsatisfactory depth reasoning without explicit height or age awareness. \cref{fig:supp_failure} (bottom) shows poor mesh estimations on challenging scenes with heavy occlusion and complex human poses, which also challenges existing \ac{sota} methods \cite{sun2024aios, multihmr2024}.

\subsection{Discussion} \label{sec:supp_discuss}

Multi-person 3D human mesh estimation is a fundamental task with broad applications. With recent one-stage \ac{sota} methods \cite{sun2024aios, multihmr2024} achieving remarkable improvements in accuracy, we further explore the potential of DETR-style pipeline by leveraging scale-adaptive tokens to encode features more efficiently. Our approach achieves superior performance with significantly lower computational cost, marking a step forward for real-time applications. With more diverse training data of high quality \ac{gt}s, we may further enhance our model's robustness and generalization capability. Additionally, our scale-adaptive tokens may be able to be plugged into other DETR-style works to improve their efficiency in the future.

\vspace{0.8em}
\noindent\textbf{Limitations.} Since our method is not age- or height-aware, it may produce larger depth estimation errors for children, as shown in \cref{fig:supp_failure} (top). In the future, this issue could be addressed by incorporating a mechanism to identify and account for children. 
Also, we currently only support body-only estimation. Since regions of human face and hands are also challenging and require a higher resolution, our method can be extended to full-body estimation in the future.

\end{document}